\documentclass[11pt,letterpaper]{berkeley}

\usepackage[all]{hypcap}

\usepackage[authoryear, round]{natbib}
\usepackage{hyperref}[citecolor=lightblue]

\hypersetup{
    colorlinks = true,
    citecolor = {blue},
}

\usepackage{microtype}
\usepackage{graphicx}
\usepackage{booktabs} %
\usepackage{float}

\usepackage{amsmath}
\usepackage{amssymb}
\usepackage{mathtools}
\usepackage{amsthm}
\usepackage{mathrsfs}
\usepackage{nicefrac}
\usepackage{dsfont}
\usepackage{enumitem}
\usepackage{float}
\usepackage{subcaption}
\usepackage{xspace}
\usepackage[capitalize,noabbrev]{cleveref}
\usepackage{wrapfig}
\usepackage{lipsum}
\usepackage{listings}

\usepackage{amsmath}
\usepackage{amssymb}
\usepackage{mathtools}
\usepackage{amsthm}
\usepackage{bbm}

\usepackage{algpseudocode}
\usepackage{setspace}

\usepackage{color}
\definecolor{deepblue}{rgb}{0,0,0.5}
\definecolor{deepred}{rgb}{0.6,0,0}
\definecolor{deepgreen}{rgb}{0,0.5,0}

\newcommand\pythonstyle{\lstset{
basicstyle=\ttfamily\footnotesize,
language=Python,
morekeywords={self, clip, exp, mse_loss, uniform_sample, concatenate, logsumexp},              %
keywordstyle=\color{deepblue},
emph={MyClass,__init__},          %
emphstyle=\color{deepred},    %
stringstyle=\color{deepgreen},
frame=single,                         %
showstringspaces=false
}}

\lstnewenvironment{python}[1][]
{
\pythonstyle
\lstset{#1}
}
{}

\newcommand\pythoninline[1]{{\pythonstyle\lstinline!#1!}}
\newcommand{\methodname}{FSPO}

\makeatletter
\def\mathcolor#1#{\@mathcolor{#1}}
\def\@mathcolor#1#2#3{%
  \protect\leavevmode
  \begingroup
    \color#1{#2}#3%
  \endgroup
}
\makeatother

\usepackage[textsize=tiny]{todonotes}
\usepackage{algorithm}
\usepackage{amssymb}

\usepackage[skins,theorems]{tcolorbox}

\tcbset{
  aibox/.style={
    width=474.18663pt,
    top=7pt,
    bottom=5pt,
    colback=blue!6!white,
    colframe=black,
    colbacktitle=black,
    enhanced,
    center,
    attach boxed title to top left={yshift=-0.1in,xshift=0.15in},
    boxed title style={boxrule=0pt,colframe=white,},
  }
}
\newtcolorbox{AIbox}[2][]{aibox,title=#2,#1}

\Crefformat{equation}{#2Eq.\;(#1)#3}

\Crefformat{figure}{#2Figure #1#3}
\Crefformat{assumption}{#2Assumption #1#3}
\Crefname{assumption}{Assumption}{Assumptions}

\usepackage{crossreftools}
\newcommand{\bDiamond}{\mathbin{\Diamond}}
\newcommand{\bLozenge}{\mathbin{\blacklozenge}}

\usepackage{dsfont}
\usepackage{nicefrac}
\usepackage{inconsolata}

\title{{\methodname: Few-Shot Optimization of Synthetic Preferences Personalizes to Real Users}}

\reportnumber{} %

\author[1,$\bDiamond$]{Anikait Singh}
\author[1,$\bDiamond$]{Sheryl Hsu}
\author[1]{Kyle Hsu}
\author[1,3]{Eric Mitchell}
\author[1]{Stefano Ermon}
\author[1]{Tatsunori Hashimoto}
\author[1,2,$\bLozenge$]{Archit Sharma}
\author[1,$\bLozenge$]{Chelsea Finn}

\affil[1]{Stanford University}
\affil[2]{Google DeepMind}
\affil[3]{OpenAI}
\affil[$\bDiamond$]{Equal Contribution}
\affil[$\bLozenge$]{Equal Advising}

\correspondingauthor{anikait@stanford.edu.\\ \textbf{Project Website:} \url{https://fewshot-preference-optimization.github.io/}}

\begin{abstract}
\vspace{-0.3cm}
Effective personalization of LLMs is critical for a broad range of user-interfacing applications such as virtual assistants and content curation. Inspired by the strong in-context capabilities of LLMs, we propose few-shot preference optimization (\methodname{}), an algorithm for LLM personalization that reframes reward modeling as a meta-learning problem. Under \methodname{}, an LLM learns to quickly infer a personalized reward function for a user via a few labeled preferences. \methodname{} also utilizes user description rationalization (RAT) to encourage better reward modeling and instruction following, recovering performance with the oracle user description. Since real-world preference data is challenging to collect at scale, we propose careful design choices to construct synthetic preference datasets for personalization, generating over 1M synthetic personalized preferences using publicly available LLMs. To successfully transfer from synthetic data to real users, we find it crucial for the data to exhibit both high diversity and coherent, self-consistent structure. We evaluate \methodname{} on personalized open-ended generation for up to 1,500 synthetic users across three domains: movie reviews, education, and open-ended question answering. We also run a controlled human study. Overall, \methodname{} achieves an 87\% Alpaca Eval winrate in generating responses that are personalized to synthetic users and a 70\% winrate with real human users in open-ended question answering. 
\end{abstract}

\begin{document}

\maketitle

\begin{flushright}
``Every story I create, creates me. I write to create myself.'' \\---Octavia E. Butler 
\end{flushright}

\vspace{-0.15cm}
\section{Introduction}  \label{sec:intro}
\vspace{-0.15cm}

As large language models (LLMs) increasingly interact with a diverse user base, it becomes important for models to generate responses that align with individual user preferences. People exhibit a wide range of preferences and beliefs shaped by their cultural background, personal experience, and individual values. These diverse preferences are present in human-annotated preference datasets; however, current preferences optimization techniques like reinforcement learning from human feedback (RLHF) largely focus on optimizing a \emph{single} model based on preferences aggregated over the entire population. This approach may neglect minority viewpoints, embed systematic biases into the model, and ultimately lead to worse performance compared to personalized models. Can we create language models that can adaptively align with the personal preferences of each user instead of the aggregated preferences of all users? 

Addressing this challenge requires a shift from modeling a singular aggregate reward function to modeling a distribution of reward functions that captures the diversity of human preferences~\citep{sorensen2024roadmappluralisticalignment, jang2023personalizedsoupspersonalizedlarge}. By doing so, we can enable personalization in language models, allowing them to generate a wide range of responses tailored to individual subpopulations. This approach not only enhances user satisfaction but also promotes inclusivity by acknowledging and respecting the varied perspectives that exist within any user base. Despite this problem's importance, to our knowledge LLM personalization has yet to be achieved for open-ended question answering with real users.

\begin{figure*}
    \centering
    \includegraphics[width=\linewidth]{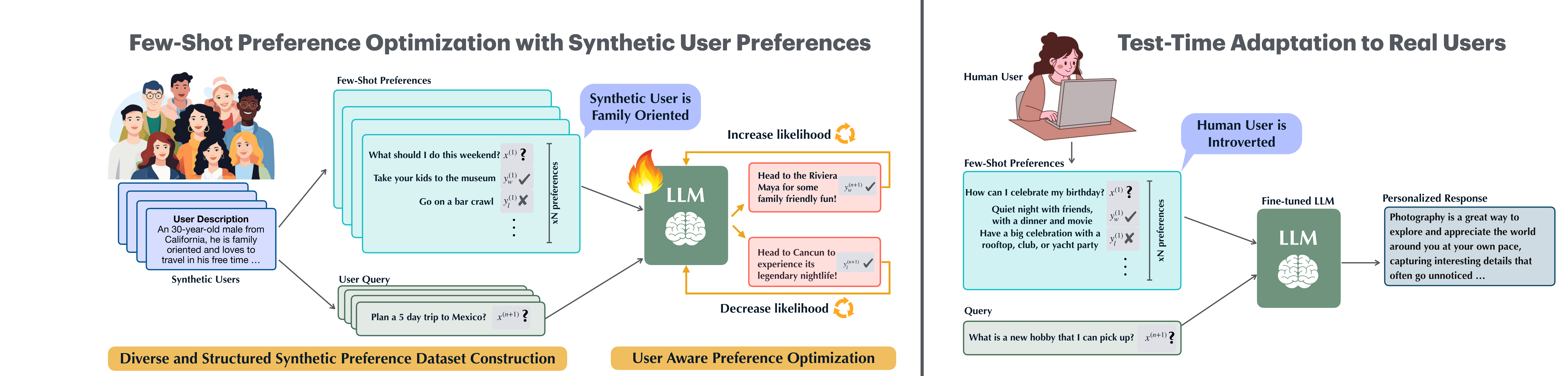}
    \caption{\footnotesize\textbf{ Overview of \methodname.} $N$ previously collected preferences are fed into the LLM along with the current query, allowing the LLM to personalize its response to the query using the past preferences. Furthermore, user description rationalization (e.g Synthetic user is family-oriented) is utilized to predict details about a user from their preferences in natural language, aiding reward modeling and text generation.}
    \vspace{-0.3cm}
    \label{fig:fspo-overview}
\end{figure*}

In this paper, we introduce few-shot preference optimization  (\methodname{}), a novel framework designed to model diverse subpopulations in preference datasets to elicit personalization in language models for open-ended question answering. At a high level, \methodname\ leverages in-context learning to adapt to new subpopulations. This adaptability is crucial for practical applications, where user preferences can be dynamic and multifaceted. Inspired by past work on black-box meta-learning for language modeling~\citep{chen2022metalearninglanguagemodelincontext, min2022metaicllearninglearncontext, yu2024metamathbootstrapmathematicalquestions}, we fine-tune the model in a meta-learning setup using preference-learning objectives such as IPO~\citep{2023arXiv231012036G}. To further improve personalized generation, we additionally propose user description rationalization (RAT), which allows the model to leverage additional inference-time compute for better reward modeling and instruction following.

Learning a model that effectively personalizes to real people requires training on a realistic, user-stratified preference dataset. One natural approach to consider is to curate such data from humans, but this is difficult and time-consuming. Instead, we propose instantiating this dataset synthetically, and present careful design decisions inspired from the meta-learning literature~\citep{hsu2019unsupervised,yin2019meta} to generate a dataset that is both diverse and structured.

To evaluate the efficacy of our approach, we construct a set of three semi-realistic domains to study personalization: (1) \textbf{Reviews}, studying the generation ability of models for reviews of movies, TV shows, and books that are consistent with a user’s writing style, (2) \textbf{Explain Like I'm X (ELIX)}: studying the generation ability of models for responses that are consistent with a user’s education level, and (3) \textbf{Roleplay}: studying the generation ability of models for responses that are consistent with a user's description, with effective transferability to a real human-study. Here we find that \methodname\ outperforms an unpersonalized model on average by 87\%. We additionally perform a controlled human study showcasing a winrate of 70\% of \methodname\ over unpersonalized models.

By addressing limitations of existing reward modeling techniques, our work paves the way for more inclusive and personalized LLMs. We believe that \methodname\ represents a significant step toward models that better serve the needs of all users, respecting the rich diversity of human preferences.
\vspace{-0.15cm}
\section{Related Work}
\vspace{-0.15cm}

\noindent \textbf{Personalized learning of preferences.}
Prior research has explored personalization through various methods. One approach is distributional alignment, which focuses on matching model outputs to broad target distributions rather than tailoring them to individual user preferences. For example, some prior work have concentrated on aligning model-generated distributions with desired statistical properties~\citep{siththaranjan2024distributionalpreferencelearningunderstanding,meister2024benchmarkingdistributionalalignmentlarge, melnyk2024distributionalpreferencealignmentllms}, yet they do not explicitly optimize for individual preference adaptation. Another strategy involves explicitly modeling a distribution of rewards~\citep{lee2024testtimealignmenthypothesisreweighting, poddar2024personalizingreinforcementlearninghuman}. However, these methods suffer from sample inefficiency during both training and inference~\citep{rafailov2023direct,2023arXiv231012036G}. Additionally, these approaches have limited evaluations: \citet{lee2024testtimealignmenthypothesisreweighting} focuses solely on reward modeling, while \citet{poddar2024personalizingreinforcementlearninghuman} tests with a very limited number of artificial users (e.g helpfulness user and honest user). Other works have investigated personalization in multiple-choice questions, such as GPO \citep{zhao2024grouppreferenceoptimizationfewshot}. Although effective in structured survey settings, these methods have not been validated for open-ended personalization tasks. Similarly, \citet{shaikh2024showdonttellaligning} explores personalization via explicit human corrections, but relying on such corrections is expensive and often impractical to scale.  Finally, several datasets exist for personalization, such as Prism~\citep{kirk2024prismalignmentdatasetparticipatory} and Persona Bench \citep{castricato2024personareproducibletestbedpluralistic}. Neither of these datasets demonstrate that policies trained on these benchmarks lead to effective personalization. Unlike these prior works which study personalization based off of human values and controversial questions, we instead study more general questions that a user may ask.

\noindent \textbf{Algorithms for preference learning.}
LLMs are typically fine-tuned via supervised next-token prediction on high-quality responses and later refined with human preference data~\citep{casper2023open,ouyang2022training}. This process can use on-policy reinforcement learning methods like REINFORCE~\citep{NIPS1999_464d828b} or PPO~\citep{2017arXiv170706347S}, which optimize a reward model with a KL constraint. Alternatively, supervised fine-tuning may be applied to a curated subset of preferred responses~\citep{dubois2024alpacafarm} or iteratively to preferred completions as in ReST~\citep{gulcehre2023reinforced}. Other methods, such as DPO~\citep{rafailov2023direct}, IPO~\citep{2023arXiv231012036G}, and KTO~\citep{HALOs2024}, learn directly from human preferences without an explicit reward model, with recent work exploring iterative preference modeling applications~\citep{2024arXiv240110020Y}.

\noindent \textbf{Black-box meta-learning.}
\methodname{} is an instance of black-box meta-learning, which has been studied in a wide range of domains spanning image classification~\citep{santoro2016oneshotlearningmemoryaugmentedneural, mishra2018simpleneuralattentivemetalearner}, language modeling~\citep{chen2022metalearninglanguagemodelincontext, min2022metaicllearninglearncontext, yu2024metamathbootstrapmathematicalquestions}, and reinforcement learning~\citep{duan2016rl, wang2016learning}. Black-box meta-learning is characterized by the processing of task contexts and queries using generic sequence operations like recurrence or self-attention, instead of specifically designed adaptation mechanisms.
\vspace{-0.15cm}
\section{Preliminaries and Notation}
\label{sec:background}
\vspace{-0.15cm}

Preference fine-tuning algorithms, such as reinforcement learning from human feedback (RLHF) and direct preference optimization (DPO), typically involve two main stages~\citep{ouyang2022training,2022arXiv220302155O}: supervised fine-tuning (SFT) and preference optimization (DPO/RLHF). First, a pre-trained model is fine-tuned on high-quality data from the target task using SFT. This process produces a reference model, denoted as $\pi_{\text{ref}}$. The purpose of this stage is to bring the responses from a particular domain in distribution with supervised learning. To further refine $\pi_{\text{ref}}$ according to human preferences, a preference dataset $\mathcal{D}_{\text{pref}} = \{(\mathbf{x}^{(i)}, \mathbf{y}_w^{(i)}, \mathbf{y}_l^{(i)})\}$ is collected. In this dataset, $\mathbf{x}^{(i)}$ represents a prompt or input context, $\mathbf{y}_w^{(i)}$ is the preferred response, and $\mathbf{y}_l^{(i)}$ is the less preferred response. These responses are typically sampled from the output distribution of $\pi_{\text{ref}}$ and are labeled based on human feedback.

Most fine-tuning pipelines assume the existence of an underlying reward function $r^*(\mathbf{x}, \cdot)$ that quantifies the quality of responses. A common approach to modeling human preferences is the Bradley-Terry (BT) model~\citep{bradleyterry1952preferences}, which expresses the probability of preferring response $\mathbf{y}_1$ over $\mathbf{y}_2$, given a prompt $\mathbf{x}$, as:
\begin{align}
\label{eq:bradley_terry}
p^*(\mathbf{y}_1 \succ \mathbf{y}_2 \mid \mathbf{x}) = \frac{e^{r^*(\mathbf{x}, \mathbf{y}_1)}}{e^{r^*(\mathbf{x}, \mathbf{y}_1)} + e^{r^*(\mathbf{x}, \mathbf{y}_2)}}
\end{align}
Here, $p^*(\mathbf{y}_1 \succ \mathbf{y}_2 \mid \mathbf{x})$ denotes the probability that $\mathbf{y}_1$ is preferred over $\mathbf{y}_2$ given $\mathbf{x}$.

The objective of preference fine-tuning is to optimize the policy $\pi_{\theta}$ to maximize the expected reward $r^*$. However, directly optimizing $r^*$ is often impractical due to model limitations or noise in reward estimation. Therefore, a reward model $r_{\phi}$ is trained to approximate $r^*$. To prevent the fine-tuned policy $\pi_{\theta}$ from deviating excessively from the reference model $\pi_{\text{ref}}$, a Kullback-Leibler (KL) divergence constraint is imposed. This leads to the following fine-tuning objective:
\begin{align}
\label{eq:rlhf_objective}
\begin{split}
\max_{\pi} \; \mathbb{E}[r^*(x,y)] - \beta\, D_\text{KL}(\pi\parallel\pi_{\text{ref}})
\end{split}
\end{align}
In this equation, the regularization term weighted by $\beta$ controls how much $\pi_{\theta}$ diverges from $\pi_{\text{ref}}$, based on the reverse KL divergence constraint. This constraint ensures that the updated policy remains close to the reference model while improving according to the reward function.

\noindent \textbf{Reward model training.} To fine-tune the large language model (LLM) policy $\pi_{\theta}(\mathbf{y} \mid \mathbf{x})$, the Bradley-Terry framework allows for either explicitly learning a reward model $r_{\phi}(\mathbf{x}, \mathbf{y})$ or directly optimizing preferences. Explicit reward models are trained using the following classification objective:
\begin{align}
\label{eq:reward_learning}
\max_{\phi}~ \mathbb{E}_{\mathcal{D}_{\text{pref}}} \left[\log \sigma \left(r_{\phi}(\mathbf{x}, \mathbf{y}_w) - r_{\phi}(\mathbf{x}, \mathbf{y}_l) \right) \right]
\end{align}
where $\sigma$ is the logistic function, used to map the difference in rewards to a probability. Alternatively, contrastive learning objectives such as Direct Preference Optimization~\citep{rafailov2023direct} and Implicit Preference Optimization~\citep{2023arXiv231012036G} utilize the policy’s log-likelihood $\log \pi_{\theta}(\mathbf{y} \mid \mathbf{x})$ as an implicit reward:
\begin{align} \label{eq:contrastive_parameterization} r_{\theta}(\mathbf{x}, \mathbf{y}) = \beta \log \big( \pi_{\theta}(\mathbf{y} \mid \mathbf{x}) / \pi_{\text{ref}}(\mathbf{y} \mid \mathbf{x}) \big) \end{align}
This approach leverages the policy’s log probabilities to represent rewards, thereby simplifying the reward learning process.
\vspace{-0.15cm}
\section{The Few-Shot Preference Optimization (\methodname) Framework}
\vspace{-0.15cm}

\begin{figure}
    \centering
    \includegraphics[width=0.8\linewidth]{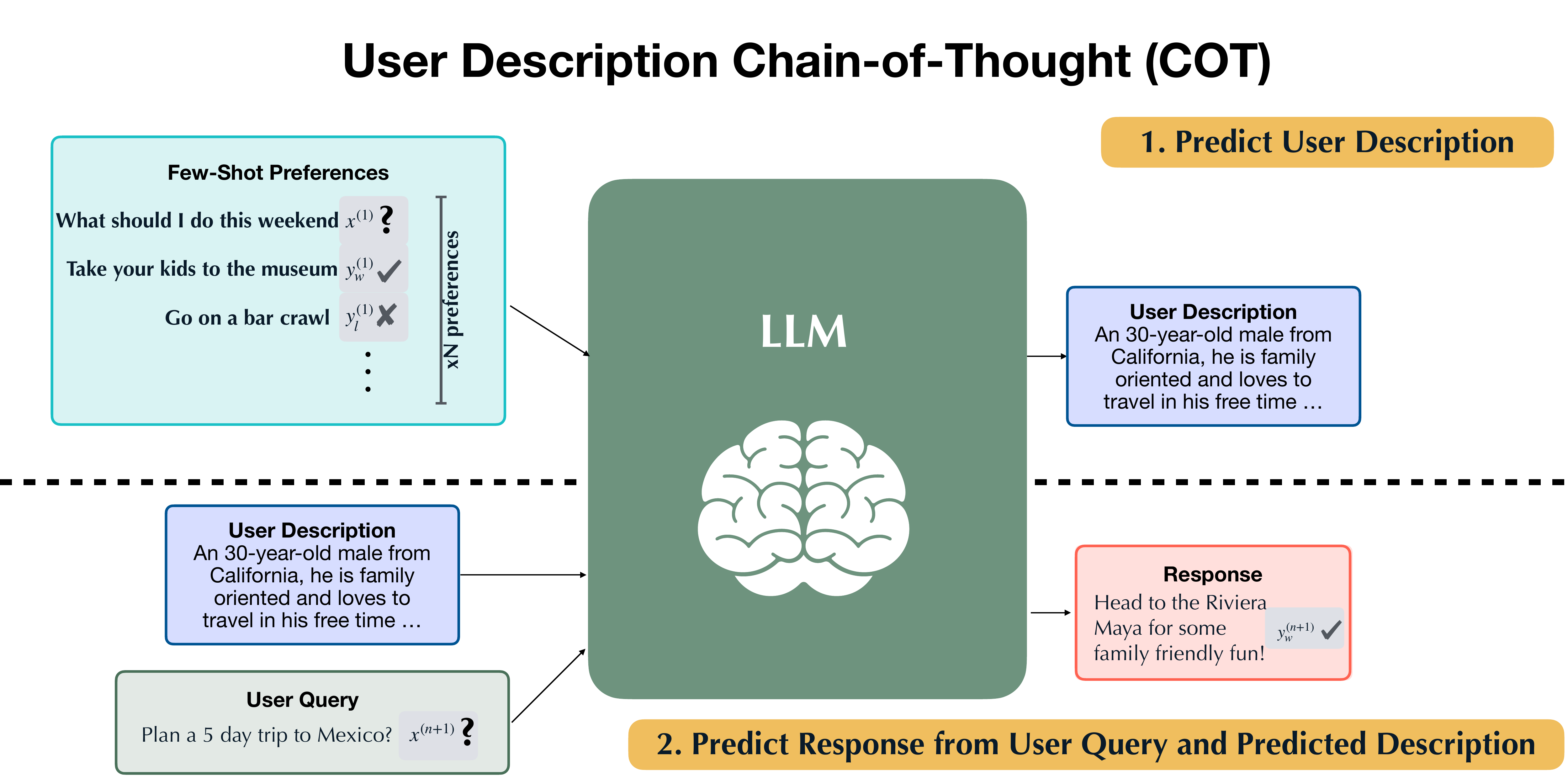}
    \caption{\footnotesize\textbf{User Description Rationalization (RAT).} Prediction is a two-stage process: first predicting a (synthetic) user description from the few-shot preferences and next predicting the response. The model is fine-tuned with a reward of how close the generated user description is to the gold user description.}
    \label{fig:persona_cot}
    \vspace{-0.4cm}
\end{figure}

\textbf{Personalization as a meta-learning problem.} Generally, for fine-tuning a model with RLHF a preference dataset of the form: $\mathcal{D}_{\text{pref}} = \{(\mathbf{x}^{(i)}, \mathbf{y}_w^{(i)}, \mathbf{y}_l^{(i)})\}$ is collected, where $x$ is a prompt, $y_w$ is a preferred response, and $y_l$ is a dispreferred response. Here, preferences from different users are aggregated to learn the preferences over a population. However, through this aggregation, individual user preferences are marginalized, leading to the model losing personalized values or beliefs due to population-based preference learning and RLHF algorithms such as DPO as seen in prior work~\citep{siththaranjan2024distributionalpreferencelearningunderstanding}.

How can we incorporate user information when learning from preference datasets? In this work, we have a weak requirement to collect scorer-ids $\mathbf{S}^{(i)}$ of each user for differentiating users that have labeled preferences in our dataset: $\mathcal{D}_{\text{pref}} = \{(\mathbf{x}^{(i)}, \mathbf{y}_w^{(i)}, \mathbf{y}_l^{(i)}, \mathbf{S}^{(i)})\}$. Now consider each user as a task instance, where the objective is to learn an effective reward function for that user using the user's set of preferences. This can be naturally instantiated as a black-box meta-learning objective, where meta-learning is done over users (also referred to as a task in meta-learning). Meta-learning should enable rapid personalization, i.e. adaptability to new users with just a few preferences.

More formally, consider that each unique user $\mathcal{S}^{(i)}$'s reward function is characterized by a set of preferences with prompt and responses $(x, y_1, y_2)$, and preference label $c$ (indicating if $y_1 \succ y_2$ or $y_1 \prec y_2$). Given a distribution over users $\mathcal{S}=P(\mathcal{S}^{(i)})$, a meta-learning objective can be derived to minimize its expected loss with respect to $\theta$ as:
\begin{align}
    \min_{\theta} \mathbb{E}_{\mathcal{S}^{(i)} \sim \mathcal{S}} \left[\mathbb{E}_{(x,y_1,y_2,c) \sim \mathcal{D}_i, \{(x,y_1,y_2,c)\}_{1}^{N} \sim \mathcal{D}_i}\left[\mathcal{L}^{\theta}_{\textit{pref}}\left(x,y_1,y_2,c \vert \{(x,y_1,y_2,c)\}_{1}^{N} \right)\right]\right]
    \label{eq:metalearn-obj}
\end{align}
where $D_i$ is a distribution over preference tuples $(x,y_1,y_2,c)$ for each user $S^{(i)}$, and $\mathcal{L}^{\theta}_{\textit{pref}}$ is a preference learning objective such as DPO~\citep{rafailov2023direct} or IPO~\citep{2023arXiv231012036G}:
\begin{align}
    \mathcal{L}^{\theta}_{\textit{pref}} = || h_{\pi_\theta}^{y_w,y_l} - (2\beta)^{-1}||^2_{2}, ~~~~
    h_{\pi_\theta}^{y_w,y_l} = \log \frac{\pi_\theta(y_w| x)}{\pi_{\textit{ref}}(y_w| x)} - \log \frac{\pi_\theta(y_l| x)}{\pi_{\textit{ref}}(y_l| x)}
    \label{eq:ipo-obj}
\end{align}
where $y_w$ and $y_l$ are the preferred and dispreferred responses (respectively) according to the responses $y_1,y_2$ and class label $c$ in the preference dataset.

Following black-box meta-learning approaches, \methodname\ receives as input a sequence of preferences $D_i^{fewshot} \sim D_i$ from a User $S^{(i)}$. This is followed by an unlabeled, held-out preference $(x,y_1,y_2) \sim \mathcal{D}_i \backslash \mathcal{D}_i^{\textit{fewshot}}$ for which it outputs its prediction $c$. To make preferences compatible with a pre-trained language model, a few-shot prompt is constructed, comprising of preferences from a user and the held-out query as seen in Figure~\ref{fig:fspo-overview}. This construction has an added benefit of leveraging a pretrained language model's capabilities for few-shot conditioning~\citep{brown2020language}, which can enable some amount of steerage/personalization. This prediction $c$ is implicitly learned by a preference optimization algorithm such as DPO~\citep{rafailov2023direct}, which parameterizes the reward model as $\beta\frac{\log \pi_{\theta}(y|x)}{\log \pi_{\textit{ref}}(y|x)}$. This parameterization enables us to leverage the advantages of preference optimization algorithms such as eliminating policy learning instabilities and computational burden of on-policy sampling, learning an effective model with a simple classification objective.

\noindent \textbf{User description rationalization (RAT).} If provided with a description of the user (potentially synthetically generated), \methodname\ can be converted to a two-step prediction problem as seen in Figure~\ref{fig:persona_cot}. In the first step, conditioned on user few-shot preferences, the user description is generated, then conditioned on the prompt, few-shot preferences, and generated user description, a response can then be generated (Example in Appendix~\ref{sec:user_desc_rat}). This prediction of the user description is an interpretable summarization of the fewshot preferences and a better representation to condition on for response generation. Similar to the rationale generated in \citet{zhang2024generativeverifiersrewardmodeling} for verifiers, the RAT prediction can be viewed as using additional inference-compute for better reward modeling. Additionally, this formulation leverages the instruction following ability of LLMs~\citep{ouyang2022training} for response generation. 

This rationalization procedure is expert-guided, fine-tuned with preference pairs over on-policy samples of a user description, where a preferred user description is one that is semantically closer to the ground-truth user description, conditioned on few-shot examples from the user. This benefits the optimization procedure twofold by (1) leveraging additional inference-compute for better reward modeling and (2) utilizing the instruction-following ability of LLMs for response generation. The instantiation of this rationalization optimization is unique, fundamentally different from COT approaches present in reasoning tasks, which use rule-based rewards to train Long-COT models for math and code reasoning. For an open-ended task, such verifiers do not exist and thus requires a different instantiation. We additionally show in Appendix A.2.1, a sample persona generated with RAT and that it qualitatively matches the underlying held-out user description, showing the efficacy of the procedure to recover characteristics about an \emph{unseen user}.

\noindent \textbf{User representation through preference labels}. From an information-theoretic perspective, the few-shot binary preferences can be seen as a $N$-bit representation of the user, representing up to $2^N$ different personas or reward functions. There are several ways to represent users: surveys, chat histories, or other forms of interaction that reveal hidden preferences. We restrict our study to such a $N$-bit user representation, as such a constrained representation can improve the performance when transferring reward models learned on synthetic personalities to real users. We defer the study of less constrained user representations to future work.

\begin{center}
\begin{minipage}{\linewidth}
    \begin{algorithm}[H]
    \caption{Overview of Few-Shot preference optimization (\methodname).}
    \begin{algorithmic}[1]
    \State \textbf{Input:} For each unique user $\mathcal{S}^{(i)}$, a dataset of preferences $\mathcal{D}:={(x,y_1,y_2,c)}_{i}$, and optionally user description $y_{\mathcal{S}^{(i)}}^{\text{+}}$, $y_{\mathcal{S}^{(i)}}^{\text{-}}$ for RAT ($+$ is preferred user description and $-$ is dispreffered user description given gold user description $y_{\mathcal{S}^{(i)}}^{\text{*}}$), $\forall i$
    \State \textbf{Output:} Learned policy $\pi_{\theta}$
    \While{not done}
    \State Sample training user $\mathcal{S}^{(i)}$ (or minibatch)
    \State Sample a subset of preferences from the user $\mathcal{D}_i^{\textit{fewshot}} \sim \mathcal{D}_i$
    \State Sample held-out preference examples $D_{i}^{\textit{heldout}} \sim \mathcal{D}_i \backslash \mathcal{D}_i^{\textit{fewshot}}$
    \If{RAT} 
    \State Use \cref{eq:metalearn-obj} and \cref{eq:ipo-obj} to predict the loss on the user descriptions $y_{\mathcal{S}^{(i)}}^{+}$ and $y_{\mathcal{S}^{(i)}}^{-}$. 
    \EndIf
    \State Conditioning on $\mathcal{D}_i^{\textit{fewshot}}$ (optionally $y_{\mathcal{S}^{(i)}}$), use \cref{eq:metalearn-obj} and \cref{eq:ipo-obj} to predict the loss on the held-out preference example $D_{i}^{\textit{heldout}}$
    \State Update learner parameters $\theta$, using gradient of loss on $D_{i}^{\textit{heldout}}$
    \EndWhile
    \State \textbf{Return} $\pi_{\theta}$
    \end{algorithmic}
    \label{alg:fspo}
    \end{algorithm}
\end{minipage}
\end{center}

We summarize \methodname\ in Algorithm~\ref{alg:fspo}. Next, we will discuss domains to study \methodname.


\section{Constructing a TestBed for Personalization}
\label{sec:domains}

To study personalization with \methodname\, we construct a benchmark across 3 domains ranging from generating personalized movie reviews (\textbf{Reviews}), generating personalized responses based off a user's education background (\textbf{ELIX}), and personalizing for general question answering (\textbf{Roleplay}). We open-source preference datasets and evaluation protocols from each of these tasks for future work looking to study personalization (sample in supplementary).

\noindent \textbf{Reviews.} The Reviews task is inspired by the IMDB dataset~\citep{maas-etal-2011-learning}, containing reviews for movies. We curate a list of popular media such as movies, TV shows, anime, and books for a language model to review. We consider two independent axes of variation for users: sentiment (positive and negative) and conciseness (concise and verbose). Here being able to pick up the user is crucial as the users from the same axes (e.g positive and negative) would have opposite preferences, making this \emph{difficult} to learn with any population based RLHF method. We also study the steerability of the model considering the axes of verbosity and sentiment in tandem (e.g positive + verbose). 

\noindent \textbf{ELIX.} The Explain Like I'm X (ELIX) task is inspired by the subreddit "Explain Like I'm 5" where users answer questions at a very basic level appropriate for a 5 year old. Here we study the ability of the model to personalize a pedagogical explanation to a user's education background. We construct two variants of the task. The first variant is \textbf{ELIX-easy} where users are one of 5 education levels (elementary school, middle school, high school, college, expert) and the goal of the task is to explain a question such as ``How are beaches formed?'' to a user of that education background. The second, more realistic variant is \textbf{ELIX-hard}, which consists of question answering at a high school to university level. Here, users may have different levels of expertise in different domains. For example, a PhD student in computer science may have a very different educational background from an undergraduate studying studying biology, allowing for preferences from diverse users (550 users). 

\noindent \textbf{Roleplay.} The Roleplay task tackles general question answering across a wide set of users, following PRISM~\citep{kirk2024prismalignmentdatasetparticipatory} and PERSONA Bench~\citep{castricato2024personareproducibletestbedpluralistic} to study personalization representative of the broad human population. We start by identifying three demographic traits (age, geographic location, and gender) that humans differ in that can lead to personalization. For each trait combination, we generate 30 personas, leading to 1,500 total personas. To more accurately model the distribution of questions, we split our questions into two categories: global and specific. Global questions are general where anyone may ask it, but specific questions revolve around a trait, for example an elderly person asking about retirement or a female asking about breast cancer screening.

One crucial detail for each task is the construction of a preference dataset that spans multiple users. But how should one construct such a dataset that is realistic and effective?

\begin{figure*}
    \centering
    \includegraphics[width=\linewidth]{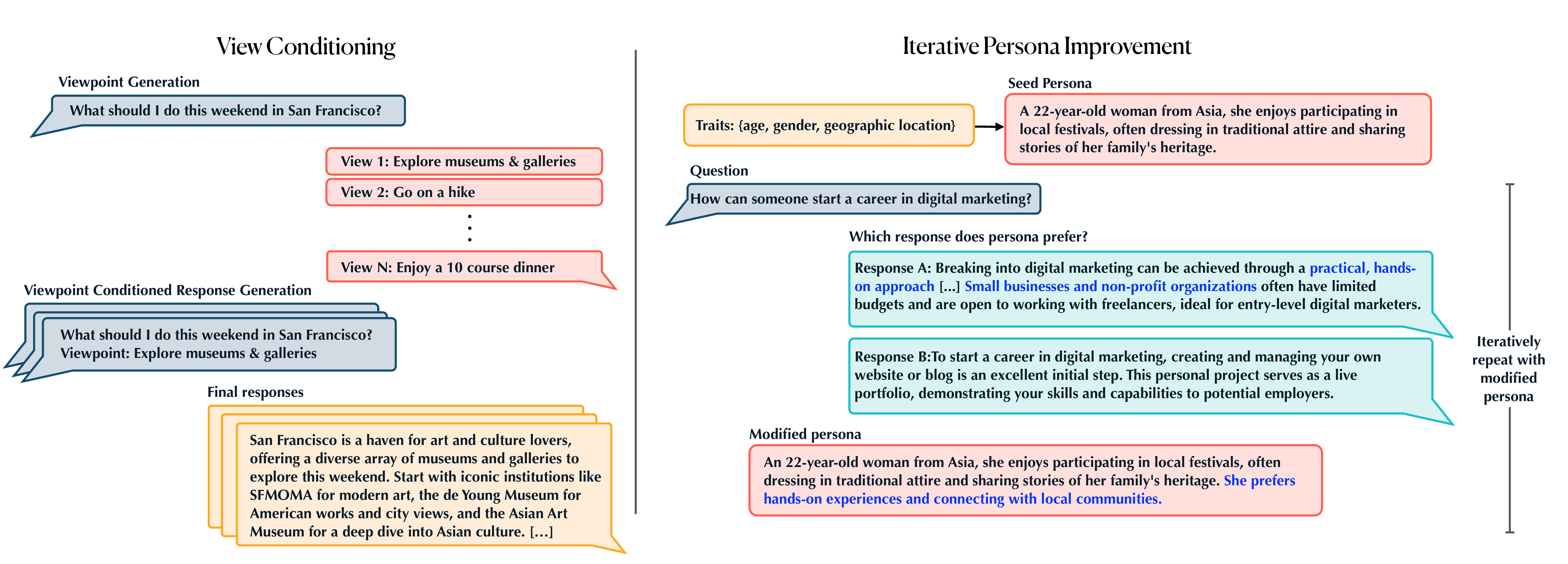}
    \caption{\footnotesize\textbf{Two key components in our synthetic data pipeline to aid with diversity and structure.} The left panel illustrates our method for increasing data diversity: we prompt a model to generate multiple viewpoints for a question and then condition our final response generation on these viewpoints. This yields greater diversity than temperature-based sampling. The right panel describes iterative persona improvement. If a seed persona is too underspecified for a clear preference, we iteratively refine its definition until it can make a robust prediction. }
    \label{fig:dom-rand}
    \vspace{-0.6cm}
\end{figure*}

\vspace{-0.15cm}
\section{Sim2Real: Synthetic Preference Data Transfers to Real Users}
\vspace{-0.15cm}
\label{sec:dom_rand}

Collecting personalized data at scale presents significant challenges, primarily due to the high cost and inherent unreliability of human annotation. Curating a diverse set of users to capture the full spectrum of real-world variability further complicates the process, often limiting the scope and representativeness of the data. Synthetically generating data using a language model~\citep{li2024syntheticdataalmostscratch,bai2022constitutionalaiharmlessnessai} is a promising alternative, since it can both reduce costly human data generation and annotation and streamline the data curation process. We note that the use of synthetic data for personalization is nuanced and amenable in many applications, as explored in Appendix~\ref{sec:synthetic_claim}. Can we generate diverse user preference data using language models in a way that transfers to real people? 

We draw inspiration from simulation-to-real transfer in non-language domains like robotics~\citep{makoviychuk2021isaacgymhighperformance} and self-driving cars~\citep{yang2023unisimneuralclosedloopsensor}, where the idea of domain randomization~\citep{tobin2018domainrandomizationgenerativemodels} has been particularly useful in enabling transfer to real environments. Domain randomization enables efficient adaptation to novel test scenarios by training models in numerous simulated environments with varied, randomized properties, enabling transfer to a held-out, real environment through interpolation.

But why is this relevant to personalization? As mentioned previously, each user can be viewed as a different ``environment'' to simulate as each user has a unique reward function that is represented by their preferences. To ensure models trained on synthetic data generalize to real human users, we employ domain randomization to simulate a diverse set of synthetic preferences. However, diversity alone isn't sufficient to learn a personalized LM. As studied in prior work~\citep{hsu2019unsupervised,yin2019meta}, it is crucial that the task distribution in meta-learning exhibits sufficient structure to rule out learning shortcuts that do not generalize. But how can we elicit both \textbf{diversity} and \textbf{structure} in our preference datasets?

\begin{figure}[t]
    \centering
    \includegraphics[width=\linewidth]{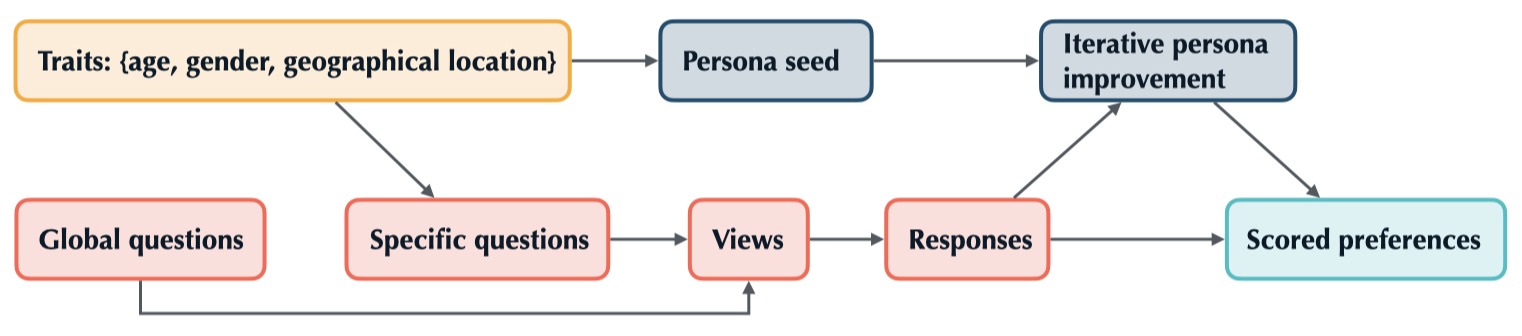}
    \vspace{-0.4cm}
    \caption{\footnotesize \textbf{Flowchart of Roleplay dataset generation:} Starting from a set of traits, a seed persona is constructed and a set of specific questions about that trait. Then responses are constructed with View-Conditioning. The seed personas are then iteratively refined to not be underspecified. Finally, the refined persona is used to score consistent preferences.}
    \label{fig:flowchart}
    \vspace{-0.8cm}
\end{figure}

\begin{wrapfigure}{r}{0.5\textwidth} 
    \centering
    \vspace{-0.2cm}
    \includegraphics[width=0.48\textwidth]{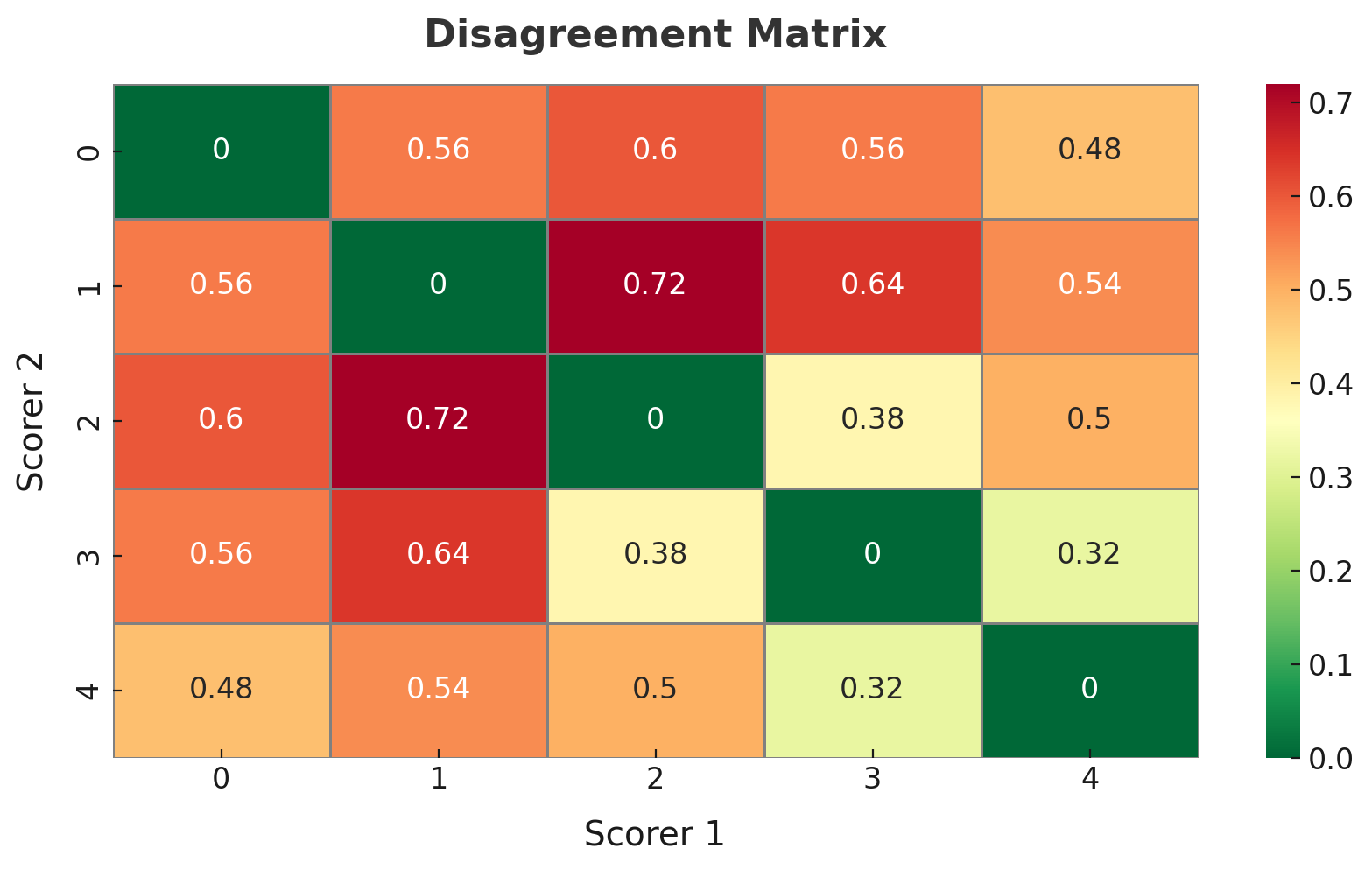}
    \caption{\footnotesize\textbf{Disagreement matrix across 5 users in Roleplay.} Here we plot the disagreement of preferences for 5 users. There is a mix of users with high and low disagreement.}
    \label{fig:disagreement}
    \vspace{-0.2cm}
\end{wrapfigure}
\noindent \textbf{Encouraging diversity.}
Diversity of data is crucial to learning a reward function that generalizes across prompts. Each domain has a slightly different generation setup as described in Section \ref{sec:domains}, but there are some general design decisions that are shared across all tasks to ensure diversity.

One source of diversity is in the questions used in the preferences. We use a variety of strategies to procure questions for the three tasks. For question selection for ELIX, we first sourced questions from human writers and then synthetically augmented the set of questions by prompting GPT-4o~\citep{openai2024gpt4ocard} with subsets of these human-generated questions. This allows us to scalably augment the human question dataset, while preserving the stylistic choices and beliefs of human writers. For the reviews dataset, we compiled a list of popular media from sites such as Goodreads, IMDb, and MyAnimeList. For the Roleplay dataset, we prompted GPT-4o to generate questions all users would ask (global) or questions only people with a specific trait would ask (specific). This allows us to have questions that are more consistent with the distribution of questions people may ask.

Additionally, having a diversity of responses is crucial for not only training the model on many viewpoints but also reward labeling, allowing for greater support over the set of possible responses for a question. To achieve diverse responses, we employ two strategies: Persona Steering~\citep{cheng-etal-2023-marked} and view conditioning (Figure~\ref{fig:dom-rand}; left). For ELIX and Reviews, we use persona steering by prompting the model with a question and asking it to generate an answer for a randomly selected persona. For Roleplay, the user description was often underspecified so responses generated with persona steering were similar. Therefore, we considered a multi-turn approach to generating a response. First, we asked the model to generate different viewpoints that may be possible for a question. Then, conditioned on each viewpoint independently, we prompted the model with the question and the viewpoint and asked it to answer the question adhering to the viewpoint presented. For example, if you consider the question, "How can I learn to cook a delicious meal?", one viewpoint here could be "watching a youtube video", better suited for a younger, more tech savvy individual, whereas viewpoints such as "using a recipe book" or "taking a cooking class" may be better for an older population or those who would have the time or money to spend on a cooking class. This allowed for more diversity in the responses and resulting preferences. 

Finally, we sampled responses from an ensemble of models with a high temperature, including those larger than the base model we fine-tuned such as Llama 3.3 70b~\citep{grattafiori2024llama} and Gemma 2 27b~\citep{gemmateam2024gemma2improvingopen}, allowing for better instruction following abilities of the fine-tuned model, than the Llama 3.2 3B we fine-tune.

\noindent \textbf{Encouraging task structure.}
Meta-learning leverages a shared latent structure across tasks to adapt to a new task quickly. The structure can be considered as similar feature representations, function families, or transition dynamics that the meta-learning algorithm can discover and leverage. For a preference dataset, this structure can be represented as the distribution of preferences across different users and is controlled by the scoring function and the distribution of responses.

One thing we controlled to enable better structure is the scoring function used to generate synthetic preferences. Firstly, we wanted to ensure consistent preference labeling. We use AI Feedback~\citep{bai2022constitutionalaiharmlessnessai} to construct this, using relative pairwise feedback for preference labels, akin to AlpacaEval~\citep{dubois2024alpacafarm}, as an alternative to absolute rubric based scoring, which we found to be noisy and inaccurate. The preference label along with being conditioned on the prompt, response, and general guidance on scoring, is now also conditioned on the scoring user description and additional scoring guidelines for user-aware preference labeling. Additionally, due to context length constraints, many responses for our preference dataset are shorter than the instruct model that we fine-tune from. Therefore, we prompt the model to ignore this bias. Furthermore, we provide each preference example to the model twice, flipping the order of the responses, and keeping filtering out responses that are not robust to order bias for both training and evaluation (win rates).

Additionally, as mentioned above, in some cases, such as with the Roleplay dataset, the user description is underspecified, leading to challenges in labeling consistent preferences. For example, if a user description does not have information about dietary preferences, inconsistency may arise for labeling preferences about that topic. For instance, in one preference pair, vegan cake recipes may be preferred but in another, steakhouses are preferred for date night. To fix this, we take an iterative process to constructing user descriptions. Firstly, we start with a seed set of user descriptions generated from the trait attributes. After generating questions and responses based on these seed descriptions, we take a set of question and response pairs. For each pair, we iteratively refine (Figure~\ref{fig:dom-rand}; right) the user description by prompting a model like GPT-4o to either label the preference pair or if the user description is insufficient, to randomly choose a preference and append information to the description so a future scorer would make the same decision. Finally, we utilize the updated user description to relabel preferences for the set of questions and responses allocated to that user with the labeling scheme above. This fix for underspecification also helps the COT prediction as predicting an underspecified user persona, can lead to ambiguous generated descriptions.

Finally, we desire structured relationships between users. To ensure this, we analyzed the disagreement (average difference of preference labels) of user's preferences across prompts to understand where users agreed and disagreed, and regenerated data if this disagreement was too high across users. By having users with some overlap, meta-learning algorithms can learn how to transfer knowledge effectively from one user to another. A sample disagreement plot for a subset of users in the Roleplay task can be found in Figure~\ref{fig:disagreement}. We outline our full dataset generation process in Figure~\ref{fig:flowchart} in the Roleplay Task, starting from just a simple set of demographic traits.


\begin{table}[ht]
    \centering
    \begin{tabular}{@{}lcc@{}}
    \toprule
    Strategy & Mean Similarity (↓) & Median Similarity (↓) \\
    \midrule
    Llama 3.2 3B Instruct, $\text{temp.}=0.3$ & 0.96 & 0.97 \\
    Llama 3.2 3B Instruct, $\text{temp.}=1.0$ & 0.94 & 0.95 \\
    Llama 3.2 3B Instruct + persona steering (ours) & 0.81 & 0.82 \\
    Llama 3.2 3B Instruct + view steering (ours) & 0.78 & 0.78 \\
    Ensemble of Models + view steering (ours) & \textbf{0.71} & \textbf{0.73} \\
    \bottomrule
    \end{tabular}
    \vspace{-0.3cm}
    \caption{Comparison of diversity-inducing strategies as evaluated under ALOE~\citep{wu2024aligning}.}
    \label{tab:similarity_scores}
    \vspace{-0.3cm}
\end{table}

\noindent\textbf{Evaluating diversity and structure.}
We evaluate our design decisions with the following vignettes. For diversity, we measure semantic similarity using the dense score from the BGE-M3 model, following ALOE~\citep{wu2024aligning}, on 100 randomly sampled prompts and 10 responses per prompt in the Roleplay task. As seen in \cref{tab:similarity_scores}, our proposed steering and ensembling mechanisms result in the base Llama 3.2 3B Instruct model exhibiting significantly reduced mean similarity. For structure, we estimate binary Shannon entropy of the preference label before and after iterative refinement. We condition on the persona and an unlabeled preference tuple (prompt and responses) and sample a preference label with a fixed temperature of $1.0$ on 100 randomly sampled prompts from the Roleplay task with 100 pairs of personas and 10 samples per prompt. We use GPT-4o as the scoring model. Iterative persona refinement causes the entropy to drop from \textbf{0.64 nats to 0.13 nats}, validating the efficacy of this approach in inducing better persona-prompt-response consistency. For further validation, we show the efficacy of scaling the size of the dataset with respect to the amount of preference data and the number of few-shot examples in Table~\ref{tab:preference_data_ablation} and Table~\ref{tab:few_shot_ablation}, showing a monotonic increase in end-to-end performance. Furthermore qualitative examples in Appendix~\ref{sec:app_dataset_eval}, showcase the diversity of viewpoints and personas as well as their alignment when scoring for structure.

\vspace{-0.15cm}
\section{Experimental Evaluation}
\vspace{-0.15cm}

\begin{table*}[t]
    \centering
    \setlength{\tabcolsep}{3pt}
    \renewcommand{\arraystretch}{1.05}
    \begin{minipage}[t]{0.48\textwidth}
        \centering
        \footnotesize
        \begin{tabular}{@{}>{\raggedright\arraybackslash}p{0.58\linewidth}cc@{}}
            \toprule
            \textbf{Method} & \textbf{Trained} & \textbf{Interpolated} \\
            \midrule
            Llama 3.2 3B Instruct & 50.0 & 50.0 \\
            \midrule
            4-shot Prompted & 66.6 & 61.9 \\
            4-shot Pref-FT  & 66.5 & 66.1 \\
            4-shot \methodname{} (Ours) & \textbf{78.4} & \textbf{71.3} \\
            \midrule
            8-shot Prompted  & 69.1 & 59.1 \\
            8-shot Pref-FT & 65.6 & 70.7 \\
            8-shot \methodname{} (Ours) & \textbf{80.4} & 
            \textbf{73.6} \\
            \midrule
            8-shot \methodname{} + RAT (Ours) &	\textbf{92.3} &	\textbf{84.6} \\
            \bottomrule
        \end{tabular}
        \caption{\footnotesize Review Winrates}
        \label{tab:winrates_reviews}
    \end{minipage}
    \hfill
    \begin{minipage}[t]{0.48\textwidth}
        \centering
        \footnotesize
        \begin{tabular}{@{}>{\raggedright\arraybackslash}p{0.76\linewidth}c@{}}
            \toprule
            \textbf{Method} & \textbf{Winrate (\%)} \\
            \midrule
            Llama 3.2 3B Instruct & 50.0 \\
            \midrule
            IPO & 72.4 \\
            Few-shot Prompting & 63.2 \\
            Few-shot Pref-FT (GPO~\citep{zhao2024grouppreferenceoptimizationfewshot}) & 62.8 \\
            RIC~\citep{yang2024rewards} & 53.3 \\
            VPL~\citep{poddar2024personalizingreinforcementlearninghuman} & 67.3 \\
            \midrule
            \methodname{} (Ours, DPO) & \textbf{81.3} \\
            \methodname{} (Ours, IPO) & \textbf{82.6} \\
            \methodname{} + RAT (Ours, IPO) & \textbf{90.3} \\
            Oracle (prompt w/ g.t. persona) & 90.9 \\
            \bottomrule
        \end{tabular}
        \caption{\footnotesize Winrates on Roleplay (1500 users)}
        \label{tab:winrates_persona}
    \end{minipage}
    
    \vspace{0.4cm}

    \begin{minipage}[t]{0.48\textwidth}
        \centering
        \footnotesize
        \begin{tabular}{@{}>{\raggedright\arraybackslash}p{0.58\linewidth}cc@{}}
            \toprule
            \textbf{Method} & \textbf{ELIX-easy} & \textbf{ELIX-hard} \\
            \midrule
            Llama 3.2 3B Instruct & 50.0 & 50.0 \\
            Few-shot Prompted & 92.4 & 81.4 \\
            Few-shot Pref-FT & 91.2 & 82.9 \\
            \methodname{} (Ours) & \textbf{97.8} & \textbf{91.8} \\
            \bottomrule
        \end{tabular}
        \caption{\footnotesize Winrates ELIX (550 users)}
        \label{tab:winrates_elix}
    \end{minipage}
    \hfill
    \begin{minipage}[t]{0.48\textwidth}
        \centering
        \footnotesize
        \begin{tabular}{@{}>{\raggedright\arraybackslash}p{0.72\linewidth}c@{}}
            \toprule
            \textbf{Baseline Method} & \textbf{Winrate (\%)} \\
            \midrule
            \methodname{} vs Base & \textbf{68.2} $\pm$ 1.93  \\
            \methodname{} vs SFT & \textbf{72.3} $\pm$ 1.34 \\
            \bottomrule
        \end{tabular}
        \caption{\footnotesize Roleplay: Human Eval Winrates}
        \label{tab:winrates_humaneval}
    \end{minipage}
    \vspace{-0.4cm}
\end{table*}

\noindent \textbf{Baselines.} We compare \methodname\ against five baselines: (1) a base model generating user-agnostic responses, (2) few-shot prompting with a base model, following~\citet{meister2024benchmarkingdistributionalalignmentlarge}, (3) few-shot supervised fine-tuning (Pref-FT) based off the maximum likelihood objective from GPO~\citep{zhao2024grouppreferenceoptimizationfewshot}, (4) prompting with an oracle user description following Persona Steering~\citep{cheng-etal-2023-marked}, and (5) Rewards-in-Context~\cite{yang2024rewards}. Specifically, for (1) we use a standard instruct model that is prompted solely with the query, resulting in unconditioned responses. For (2) and (3), the base instruct model is provided with the same few-shot personalization examples as in \methodname, but (2) zero-shot predicts the preferred response and (3) is optimized with SFT to increase the likelihood on the preferred response. In (4), the base model is prompted with the oracle, ground truth user description, representing an upper bound on \methodname’s performance.

\noindent \textbf{Synthetic winrates.} 
We first generate automated win rates using the modified AlpacaEval procedure from Section~\ref{sec:dom_rand}. In the ELIX task in Table~\ref{tab:winrates_elix}, we study two levels of difficulty (easy, hard), where we find a consistent improvement of \methodname\ over baselines. Next, in Table~\ref{tab:winrates_reviews} for the Review task, on both Trained and Interpolated Users, \methodname\ allows for better performance on held-out questions. Finally, in Table~\ref{tab:winrates_persona}, we study Roleplay, scaling to 1500 real users, seeing a win rate of 82.6\% on both held-out users and questions. Also, RAT closes the gap to the oracle response, effectively recovering the ground-truth user description. In \cref{sec:app_samples}, sample generations from \methodname\, show effective personalization to the oracle user description. Given this result, can we personalize to real people?

\noindent \textbf{Preliminary human study.} We evaluate our model trained on the Roleplay task by personalizing responses for \emph{real human participants}. We build a data collection app (Figure~\ref{fig:human_app}), interacting with a user in two stages. First, we ask participants to label preference pairs, used as the few-shot examples in \methodname. Then, for held out questions, we show a user a set of two responses: (1) a response from \methodname\ personalized based on their preferences and (2) a baseline response. Prolific is used to recruit a diverse set of study participants, evenly split across genders and continents, corresponding to the traits used to construct user descriptions. Question and response order is randomized to remove confounding factors. We evaluate with 50 users and 11 questions. As seen in Table~\ref{tab:winrates_humaneval}, we find that \methodname\ has a 68\% win rate over the Base model and a 72\% win rate over an SFT model trained on diverse viewpoints from the preference dataset. To assess statistical significance, we performed a one-sided binomial test. Here, the null hypothesis is that the probability of success is less than or equal to 50\%, (ie, that our model is no better than the baseline) and the alternative hypothesis is that the probability is greater than 50\%. The resulting p-value is 5.65e-09, so we reject the null hypothesis at any conventional significance level. We also validate \methodname{} on PRISM (Appendix~\ref{sec:prism_study}), a preference dataset on value based alignment from the community, showcasing benefits beyond our constructed datasets on real human users.


\vspace{-0.25cm}
\section{Discussion and Conclusion}
\vspace{-0.25cm}

We introduce \methodname, a novel framework for eliciting personalization in language models for open-ended question answering that models a distribution of reward functions to capture diverse human preferences. Our approach leverages meta-learning for rapid adaptation to each user, addressing limitations of conventional reward modeling techniques that learn from aggregated preferences. Through rigorous evaluation in 3 domains, we demonstrate that \methodname's generations are consistent with user context and preferred by real human users. Our findings also underscore the importance of diversity and structure in synthetic personalized preference datasets to bridge the Sim2Real gap. Overall, \methodname\ is a step towards developing more inclusive, user-centric language models.

\vspace{-0.15cm}
\section{Ethics Statement}
\vspace{-0.15cm}
While \methodname\ improves inclusivity by modeling diverse preferences, the risk of reinforcing user biases (echo chambers) or inadvertently amplifying harmful viewpoints requires careful scrutiny. Future work should explore mechanisms to balance personalization with ethical safeguards, ensuring that models remain aligned with fairness principles while respecting user individuality. Note, we choose to omit value-based personalization in the experiments as explored in works such as PRISM and Persona, instead focusing on the recommendation style of preferences such as travel preferences, where potential amplification of biases would be benign, having a limited effect on marginalizing particular subpopulations. Thus, this potential issue is a concern about using the algorithm in political or value-based contexts, not something that has arisen in the fine-tuned model. That being said, we do not explicitly mitigate this, which we leave to future work. Here, approaches such as Persona Vectors, recently released by Anthropic, can potentially be paired with an approach like FSPO to mitigate such biases in the training. We wish to emphasize clearly that our human study involves no collection of identifiable information and is strictly non-longitudinal, involving harmless, recommender-style questions. Under the criteria for Non-Medical IRBs, our study explicitly falls within the exemption specified by 45 CFR 46.104(d). Previous guidance received from our institutional IRB also confirms exemption status for such survey-based studies. Additionally, no such IRB was required in prior work including Direct Preference Optimization (DPO), AlpacaFarm, Chatbot Arena, and Persona, for nearly identical user study formulations. Thus, we strongly assert that formal IRB approval is unnecessary for our work. We additionally utilized LLMs such as GPT5/Gemini for minor rewritings of different sections throughout the paper for better readability.

\vspace{-0.2cm}
\section{Acknowledgements}
\vspace{-0.2cm}

We would like to thank Yuejiang Liu, Aviral Kumar, Rafael Rafeliov, Violet Xiang, Yoonho Lee, Meihua Dang, and others in Stanford IRIS Lab and Stanford Hashimoto Group for discussions and feedback. We would also like to thank Foundry for providing compute. AS gratefully acknowledges the support of the NSF Graduate Research Fellowship Program and Toyota Research Institute (TRI) for funding, compute and API credits. TH was supported by a HAI seed grant, gifts from Open Philanthropy, Amazon, Schmidt Sciences, the Tianqiao and Chrissy Chen Foundation and a grant under the NSF CAREER IIS-2338866. CF was supported by Schmidt Sciences.
\clearpage
\bibliography{iclr2024_conference}

@inproceedings{hsu2019unsupervised,
  title={Unsupervised learning via meta-learning},
  author={Hsu, Kyle and Levine, Sergey and Finn, Chelsea},
  booktitle={International Conference on Learning Representations},
  year={2019}
}

@article{yin2019meta,
  title={Meta-learning without memorization},
  author={Yin, Mingzhang and Tucker, George and Zhou, Mingyuan and Levine, Sergey and Finn, Chelsea},
  journal={arXiv preprint arXiv:1912.03820},
  year={2019}
}

@article{duan2016rl,
  title={Fast reinforcement learning via slow reinforcement learning},
  author={Duan, Yan and Schulman, John and Chen, Xi and Bartlett, Peter L and Sutskever, Ilya and Abbeel, Pieter},
  journal={arXiv preprint arXiv:1611.02779},
  year={2016}
}

@article{wang2016learning,
  title={Learning to reinforcement learn},
  author={Wang, Jane X and Kurth-Nelson, Zeb and Tirumala, Dhruva and Soyer, Hubert and Leibo, Joel Z and Munos, Remi and Blundell, Charles and Kumaran, Dharshan and Botvinick, Matt},
  journal={arXiv preprint arXiv:1611.05763},
  year={2016}
}

@ARTICLE{2024arXiv240110020Y,
       author = {{Yuan}, Weizhe and {Yuanzhe Pang}, Richard and {Cho}, Kyunghyun and {Li}, Xian and {Sukhbaatar}, Sainbayar and {Xu}, Jing and {Weston}, Jason},
        title = "{Self-Rewarding Language Models}",
      journal = {arXiv e-prints},
         year = 2024,
        month = jan,
          eid = {arXiv:2401.10020},
        pages = {arXiv:2401.10020},
          doi = {10.48550/arXiv.2401.10020},
archivePrefix = {arXiv},
       eprint = {2401.10020},
 primaryClass = {cs.CL},
       adsurl = {https://ui.adsabs.harvard.edu/abs/2024arXiv240110020Y}
}

@article{rafailov2023direct,
  title={Direct preference optimization: Your language model is secretly a reward model},
  author={Rafailov, Rafael and Sharma, Archit and Mitchell, Eric and Ermon, Stefano and Manning, Christopher D and Finn, Chelsea},
  journal={arXiv preprint arXiv:2305.18290},
  year={2023}
}

@misc{ouyang2022training,
      title={Training language models to follow instructions with human feedback}, 
      author={Long Ouyang and Jeff Wu and Xu Jiang and Diogo Almeida and Carroll L. Wainwright and Pamela Mishkin and Chong Zhang and Sandhini Agarwal and Katarina Slama and Alex Ray and John Schulman and Jacob Hilton and Fraser Kelton and Luke Miller and Maddie Simens and Amanda Askell and Peter Welinder and Paul Christiano and Jan Leike and Ryan Lowe},
      year={2022},
      eprint={2203.02155},
      archivePrefix={arXiv},
      primaryClass={cs.CL}
}

@article{bradleyterry1952preferences,
 ISSN = {00063444},
 URL = {http://www.jstor.org/stable/2334029},
 author = {Ralph Allan Bradley and Milton E. Terry},
 journal = {Biometrika},
 number = {3/4},
 pages = {324--345},
 publisher = {[Oxford University Press, Biometrika Trust]},
 title = {Rank Analysis of Incomplete Block Designs: I. The Method of Paired Comparisons},
 urldate = {2024-01-25},
 volume = {39},
 year = {1952}
}

@article{
casper2023open,
title={Open Problems and Fundamental Limitations of Reinforcement Learning from Human Feedback},
author={Stephen Casper and Xander Davies and Claudia Shi and Thomas Krendl Gilbert and J{\'e}r{\'e}my Scheurer and Javier Rando and Rachel Freedman and Tomasz Korbak and David Lindner and Pedro Freire and Tony Tong Wang and Samuel Marks and Charbel-Raphael Segerie and Micah Carroll and Andi Peng and Phillip Christoffersen and Mehul Damani and Stewart Slocum and Usman Anwar and Anand Siththaranjan and Max Nadeau and Eric J Michaud and Jacob Pfau and Dmitrii Krasheninnikov and Xin Chen and Lauro Langosco and Peter Hase and Erdem Biyik and Anca Dragan and David Krueger and Dorsa Sadigh and Dylan Hadfield-Menell},
journal={Transactions on Machine Learning Research},
issn={2835-8856},
year={2023},
url={https://openreview.net/forum?id=bx24KpJ4Eb},
note={Survey Certification}
}

@article{gulcehre2023reinforced,
  title={Reinforced self-training (rest) for language modeling},
  author={Gulcehre, Caglar and Paine, Tom Le and Srinivasan, Srivatsan and Konyushkova, Ksenia and Weerts, Lotte and Sharma, Abhishek and Siddhant, Aditya and Ahern, Alex and Wang, Miaosen and Gu, Chenjie and others},
  journal={arXiv preprint arXiv:2308.08998},
  year={2023}
}

@misc{dubois2024alpacafarm,
      title={AlpacaFarm: A Simulation Framework for Methods that Learn from Human Feedback}, 
      author={Yann Dubois and Xuechen Li and Rohan Taori and Tianyi Zhang and Ishaan Gulrajani and Jimmy Ba and Carlos Guestrin and Percy Liang and Tatsunori B. Hashimoto},
      year={2024},
      eprint={2305.14387},
      archivePrefix={arXiv},
      primaryClass={cs.LG}
}

@inproceedings{NIPS1999_464d828b,
 author = {Sutton, Richard S and McAllester, David and Singh, Satinder and Mansour, Yishay},
 booktitle = {Advances in Neural Information Processing Systems},
 editor = {S. Solla and T. Leen and K. M\"{u}ller},
 pages = {},
 publisher = {MIT Press},
 title = {Policy Gradient Methods for Reinforcement Learning with Function Approximation},
 url = {https://proceedings.neurips.cc/paper_files/paper/1999/file/464d828b85b0bed98e80ade0a5c43b0f-Paper.pdf},
 volume = {12},
 year = {1999}
}

@ARTICLE{2017arXiv170706347S,
       author = {{Schulman}, John and {Wolski}, Filip and {Dhariwal}, Prafulla and {Radford}, Alec and {Klimov}, Oleg},
        title = "{Proximal Policy Optimization Algorithms}",
      journal = {arXiv e-prints},
         year = 2017,
        month = jul,
          eid = {arXiv:1707.06347},
        pages = {arXiv:1707.06347},
          doi = {10.48550/arXiv.1707.06347},
archivePrefix = {arXiv},
       eprint = {1707.06347},
 primaryClass = {cs.LG},
       adsurl = {https://ui.adsabs.harvard.edu/abs/2017arXiv170706347S}
}

@ARTICLE{2022arXiv220302155O,
       author = {{Ouyang}, Long and {Wu}, Jeff and {Jiang}, Xu and {Almeida}, Diogo and {Wainwright}, Carroll L. and {Mishkin}, Pamela and {Zhang}, Chong and {Agarwal}, Sandhini and {Slama}, Katarina and {Ray}, Alex and {Schulman}, John and {Hilton}, Jacob and {Kelton}, Fraser and {Miller}, Luke and {Simens}, Maddie and {Askell}, Amanda and {Welinder}, Peter and {Christiano}, Paul and {Leike}, Jan and {Lowe}, Ryan},
        title = "{Training language models to follow instructions with human feedback}",
      journal = {arXiv e-prints},
         year = 2022,
        month = mar,
          eid = {arXiv:2203.02155},
        pages = {arXiv:2203.02155},
          doi = {10.48550/arXiv.2203.02155},
archivePrefix = {arXiv},
       eprint = {2203.02155},
 primaryClass = {cs.CL},
       adsurl = {https://ui.adsabs.harvard.edu/abs/2022arXiv220302155O}
}

@ARTICLE{2023arXiv231012036G,
       author = {{Gheshlaghi Azar}, Mohammad and {Rowland}, Mark and {Piot}, Bilal and {Guo}, Daniel and {Calandriello}, Daniele and {Valko}, Michal and {Munos}, R{\'e}mi},
        title = "{A General Theoretical Paradigm to Understand Learning from Human Preferences}",
      journal = {arXiv e-prints},
         year = 2023,
        month = oct,
          eid = {arXiv:2310.12036},
        pages = {arXiv:2310.12036},
          doi = {10.48550/arXiv.2310.12036},
archivePrefix = {arXiv},
       eprint = {2310.12036},
 primaryClass = {cs.AI},
       adsurl = {https://ui.adsabs.harvard.edu/abs/2023arXiv231012036G}
}

@misc{HALOs2024,
  author = {{ContextualAI}},
  title = {Human-Centered Loss Functions (HALOs)},
  year = {2024},
  publisher = {GitHub},
  journal = {GitHub repository},
  url = {https://github.com/ContextualAI/HALOs}
}

@misc{brown2020language,
      title={Language Models are Few-Shot Learners}, 
      author={Tom B. Brown and Benjamin Mann and Nick Ryder and Melanie Subbiah and Jared Kaplan and Prafulla Dhariwal and Arvind Neelakantan and Pranav Shyam and Girish Sastry and Amanda Askell and Sandhini Agarwal and Ariel Herbert-Voss and Gretchen Krueger and Tom Henighan and Rewon Child and Aditya Ramesh and Daniel M. Ziegler and Jeffrey Wu and Clemens Winter and Christopher Hesse and Mark Chen and Eric Sigler and Mateusz Litwin and Scott Gray and Benjamin Chess and Jack Clark and Christopher Berner and Sam McCandlish and Alec Radford and Ilya Sutskever and Dario Amodei},
      year={2020},
      eprint={2005.14165},
      archivePrefix={arXiv},
      primaryClass={cs.CL}
}

@misc{santoro2016oneshotlearningmemoryaugmentedneural,
      title={One-shot Learning with Memory-Augmented Neural Networks}, 
      author={Adam Santoro and Sergey Bartunov and Matthew Botvinick and Daan Wierstra and Timothy Lillicrap},
      year={2016},
      eprint={1605.06065},
      archivePrefix={arXiv},
      primaryClass={cs.LG},
      url={https://arxiv.org/abs/1605.06065}, 
}

@misc{mishra2018simpleneuralattentivemetalearner,
      title={A Simple Neural Attentive Meta-Learner}, 
      author={Nikhil Mishra and Mostafa Rohaninejad and Xi Chen and Pieter Abbeel},
      year={2018},
      eprint={1707.03141},
      archivePrefix={arXiv},
      primaryClass={cs.AI},
      url={https://arxiv.org/abs/1707.03141}, 
}

@inproceedings{maas-etal-2011-learning,
    title = "Learning Word Vectors for Sentiment Analysis",
    author = "Maas, Andrew L.  and
      Daly, Raymond E.  and
      Pham, Peter T.  and
      Huang, Dan  and
      Ng, Andrew Y.  and
      Potts, Christopher",
    editor = "Lin, Dekang  and
      Matsumoto, Yuji  and
      Mihalcea, Rada",
    booktitle = "Proceedings of the 49th Annual Meeting of the Association for Computational Linguistics: Human Language Technologies",
    month = jun,
    year = "2011",
    address = "Portland, Oregon, USA",
    publisher = "Association for Computational Linguistics",
    url = "https://aclanthology.org/P11-1015/",
    pages = "142--150"
}

@misc{kirk2024prismalignmentdatasetparticipatory,
      title={The PRISM Alignment Dataset: What Participatory, Representative and Individualised Human Feedback Reveals About the Subjective and Multicultural Alignment of Large Language Models}, 
      author={Hannah Rose Kirk and Alexander Whitefield and Paul Röttger and Andrew Bean and Katerina Margatina and Juan Ciro and Rafael Mosquera and Max Bartolo and Adina Williams and He He and Bertie Vidgen and Scott A. Hale},
      year={2024},
      eprint={2404.16019},
      archivePrefix={arXiv},
      primaryClass={cs.CL},
      url={https://arxiv.org/abs/2404.16019}, 
}

@misc{castricato2024personareproducibletestbedpluralistic,
      title={PERSONA: A Reproducible Testbed for Pluralistic Alignment}, 
      author={Louis Castricato and Nathan Lile and Rafael Rafailov and Jan-Philipp Fränken and Chelsea Finn},
      year={2024},
      eprint={2407.17387},
      archivePrefix={arXiv},
      primaryClass={cs.CL},
      url={https://arxiv.org/abs/2407.17387}, 
}

@misc{gemmateam2024gemma2improvingopen,
      title={Gemma 2: Improving Open Language Models at a Practical Size}, 
      author={Gemma Team and Morgane Riviere and Shreya Pathak and Pier Giuseppe Sessa and Cassidy Hardin and Surya Bhupatiraju and Léonard Hussenot and Thomas Mesnard and Bobak Shahriari and Alexandre Ramé and Johan Ferret and Peter Liu and Pouya Tafti and Abe Friesen and Michelle Casbon and Sabela Ramos and Ravin Kumar and Charline Le Lan and Sammy Jerome and Anton Tsitsulin and Nino Vieillard and Piotr Stanczyk and Sertan Girgin and Nikola Momchev and Matt Hoffman and Shantanu Thakoor and Jean-Bastien Grill and Behnam Neyshabur and Olivier Bachem and Alanna Walton and Aliaksei Severyn and Alicia Parrish and Aliya Ahmad and Allen Hutchison and Alvin Abdagic and Amanda Carl and Amy Shen and Andy Brock and Andy Coenen and Anthony Laforge and Antonia Paterson and Ben Bastian and Bilal Piot and Bo Wu and Brandon Royal and Charlie Chen and Chintu Kumar and Chris Perry and Chris Welty and Christopher A. Choquette-Choo and Danila Sinopalnikov and David Weinberger and Dimple Vijaykumar and Dominika Rogozińska and Dustin Herbison and Elisa Bandy and Emma Wang and Eric Noland and Erica Moreira and Evan Senter and Evgenii Eltyshev and Francesco Visin and Gabriel Rasskin and Gary Wei and Glenn Cameron and Gus Martins and Hadi Hashemi and Hanna Klimczak-Plucińska and Harleen Batra and Harsh Dhand and Ivan Nardini and Jacinda Mein and Jack Zhou and James Svensson and Jeff Stanway and Jetha Chan and Jin Peng Zhou and Joana Carrasqueira and Joana Iljazi and Jocelyn Becker and Joe Fernandez and Joost van Amersfoort and Josh Gordon and Josh Lipschultz and Josh Newlan and Ju-yeong Ji and Kareem Mohamed and Kartikeya Badola and Kat Black and Katie Millican and Keelin McDonell and Kelvin Nguyen and Kiranbir Sodhia and Kish Greene and Lars Lowe Sjoesund and Lauren Usui and Laurent Sifre and Lena Heuermann and Leticia Lago and Lilly McNealus and Livio Baldini Soares and Logan Kilpatrick and Lucas Dixon and Luciano Martins and Machel Reid and Manvinder Singh and Mark Iverson and Martin Görner and Mat Velloso and Mateo Wirth and Matt Davidow and Matt Miller and Matthew Rahtz and Matthew Watson and Meg Risdal and Mehran Kazemi and Michael Moynihan and Ming Zhang and Minsuk Kahng and Minwoo Park and Mofi Rahman and Mohit Khatwani and Natalie Dao and Nenshad Bardoliwalla and Nesh Devanathan and Neta Dumai and Nilay Chauhan and Oscar Wahltinez and Pankil Botarda and Parker Barnes and Paul Barham and Paul Michel and Pengchong Jin and Petko Georgiev and Phil Culliton and Pradeep Kuppala and Ramona Comanescu and Ramona Merhej and Reena Jana and Reza Ardeshir Rokni and Rishabh Agarwal and Ryan Mullins and Samaneh Saadat and Sara Mc Carthy and Sarah Cogan and Sarah Perrin and Sébastien M. R. Arnold and Sebastian Krause and Shengyang Dai and Shruti Garg and Shruti Sheth and Sue Ronstrom and Susan Chan and Timothy Jordan and Ting Yu and Tom Eccles and Tom Hennigan and Tomas Kocisky and Tulsee Doshi and Vihan Jain and Vikas Yadav and Vilobh Meshram and Vishal Dharmadhikari and Warren Barkley and Wei Wei and Wenming Ye and Woohyun Han and Woosuk Kwon and Xiang Xu and Zhe Shen and Zhitao Gong and Zichuan Wei and Victor Cotruta and Phoebe Kirk and Anand Rao and Minh Giang and Ludovic Peran and Tris Warkentin and Eli Collins and Joelle Barral and Zoubin Ghahramani and Raia Hadsell and D. Sculley and Jeanine Banks and Anca Dragan and Slav Petrov and Oriol Vinyals and Jeff Dean and Demis Hassabis and Koray Kavukcuoglu and Clement Farabet and Elena Buchatskaya and Sebastian Borgeaud and Noah Fiedel and Armand Joulin and Kathleen Kenealy and Robert Dadashi and Alek Andreev},
      year={2024},
      eprint={2408.00118},
      archivePrefix={arXiv},
      primaryClass={cs.CL},
      url={https://arxiv.org/abs/2408.00118}, 
}

@misc{openai2024gpt4ocard,
      title={GPT-4o System Card}, 
      author={OpenAI and : and Aaron Hurst and Adam Lerer and Adam P. Goucher and Adam Perelman and Aditya Ramesh and Aidan Clark and AJ Ostrow and Akila Welihinda and Alan Hayes and Alec Radford and Aleksander Mądry and Alex Baker-Whitcomb and Alex Beutel and Alex Borzunov and Alex Carney and Alex Chow and Alex Kirillov and Alex Nichol and Alex Paino and Alex Renzin and Alex Tachard Passos and Alexander Kirillov and Alexi Christakis and Alexis Conneau and Ali Kamali and Allan Jabri and Allison Moyer and Allison Tam and Amadou Crookes and Amin Tootoochian and Amin Tootoonchian and Ananya Kumar and Andrea Vallone and Andrej Karpathy and Andrew Braunstein and Andrew Cann and Andrew Codispoti and Andrew Galu and Andrew Kondrich and Andrew Tulloch and Andrey Mishchenko and Angela Baek and Angela Jiang and Antoine Pelisse and Antonia Woodford and Anuj Gosalia and Arka Dhar and Ashley Pantuliano and Avi Nayak and Avital Oliver and Barret Zoph and Behrooz Ghorbani and Ben Leimberger and Ben Rossen and Ben Sokolowsky and Ben Wang and Benjamin Zweig and Beth Hoover and Blake Samic and Bob McGrew and Bobby Spero and Bogo Giertler and Bowen Cheng and Brad Lightcap and Brandon Walkin and Brendan Quinn and Brian Guarraci and Brian Hsu and Bright Kellogg and Brydon Eastman and Camillo Lugaresi and Carroll Wainwright and Cary Bassin and Cary Hudson and Casey Chu and Chad Nelson and Chak Li and Chan Jun Shern and Channing Conger and Charlotte Barette and Chelsea Voss and Chen Ding and Cheng Lu and Chong Zhang and Chris Beaumont and Chris Hallacy and Chris Koch and Christian Gibson and Christina Kim and Christine Choi and Christine McLeavey and Christopher Hesse and Claudia Fischer and Clemens Winter and Coley Czarnecki and Colin Jarvis and Colin Wei and Constantin Koumouzelis and Dane Sherburn and Daniel Kappler and Daniel Levin and Daniel Levy and David Carr and David Farhi and David Mely and David Robinson and David Sasaki and Denny Jin and Dev Valladares and Dimitris Tsipras and Doug Li and Duc Phong Nguyen and Duncan Findlay and Edede Oiwoh and Edmund Wong and Ehsan Asdar and Elizabeth Proehl and Elizabeth Yang and Eric Antonow and Eric Kramer and Eric Peterson and Eric Sigler and Eric Wallace and Eugene Brevdo and Evan Mays and Farzad Khorasani and Felipe Petroski Such and Filippo Raso and Francis Zhang and Fred von Lohmann and Freddie Sulit and Gabriel Goh and Gene Oden and Geoff Salmon and Giulio Starace and Greg Brockman and Hadi Salman and Haiming Bao and Haitang Hu and Hannah Wong and Haoyu Wang and Heather Schmidt and Heather Whitney and Heewoo Jun and Hendrik Kirchner and Henrique Ponde de Oliveira Pinto and Hongyu Ren and Huiwen Chang and Hyung Won Chung and Ian Kivlichan and Ian O'Connell and Ian O'Connell and Ian Osband and Ian Silber and Ian Sohl and Ibrahim Okuyucu and Ikai Lan and Ilya Kostrikov and Ilya Sutskever and Ingmar Kanitscheider and Ishaan Gulrajani and Jacob Coxon and Jacob Menick and Jakub Pachocki and James Aung and James Betker and James Crooks and James Lennon and Jamie Kiros and Jan Leike and Jane Park and Jason Kwon and Jason Phang and Jason Teplitz and Jason Wei and Jason Wolfe and Jay Chen and Jeff Harris and Jenia Varavva and Jessica Gan Lee and Jessica Shieh and Ji Lin and Jiahui Yu and Jiayi Weng and Jie Tang and Jieqi Yu and Joanne Jang and Joaquin Quinonero Candela and Joe Beutler and Joe Landers and Joel Parish and Johannes Heidecke and John Schulman and Jonathan Lachman and Jonathan McKay and Jonathan Uesato and Jonathan Ward and Jong Wook Kim and Joost Huizinga and Jordan Sitkin and Jos Kraaijeveld and Josh Gross and Josh Kaplan and Josh Snyder and Joshua Achiam and Joy Jiao and Joyce Lee and Juntang Zhuang and Justyn Harriman and Kai Fricke and Kai Hayashi and Karan Singhal and Katy Shi and Kavin Karthik and Kayla Wood and Kendra Rimbach and Kenny Hsu and Kenny Nguyen and Keren Gu-Lemberg and Kevin Button and Kevin Liu and Kiel Howe and Krithika Muthukumar and Kyle Luther and Lama Ahmad and Larry Kai and Lauren Itow and Lauren Workman and Leher Pathak and Leo Chen and Li Jing and Lia Guy and Liam Fedus and Liang Zhou and Lien Mamitsuka and Lilian Weng and Lindsay McCallum and Lindsey Held and Long Ouyang and Louis Feuvrier and Lu Zhang and Lukas Kondraciuk and Lukasz Kaiser and Luke Hewitt and Luke Metz and Lyric Doshi and Mada Aflak and Maddie Simens and Madelaine Boyd and Madeleine Thompson and Marat Dukhan and Mark Chen and Mark Gray and Mark Hudnall and Marvin Zhang and Marwan Aljubeh and Mateusz Litwin and Matthew Zeng and Max Johnson and Maya Shetty and Mayank Gupta and Meghan Shah and Mehmet Yatbaz and Meng Jia Yang and Mengchao Zhong and Mia Glaese and Mianna Chen and Michael Janner and Michael Lampe and Michael Petrov and Michael Wu and Michele Wang and Michelle Fradin and Michelle Pokrass and Miguel Castro and Miguel Oom Temudo de Castro and Mikhail Pavlov and Miles Brundage and Miles Wang and Minal Khan and Mira Murati and Mo Bavarian and Molly Lin and Murat Yesildal and Nacho Soto and Natalia Gimelshein and Natalie Cone and Natalie Staudacher and Natalie Summers and Natan LaFontaine and Neil Chowdhury and Nick Ryder and Nick Stathas and Nick Turley and Nik Tezak and Niko Felix and Nithanth Kudige and Nitish Keskar and Noah Deutsch and Noel Bundick and Nora Puckett and Ofir Nachum and Ola Okelola and Oleg Boiko and Oleg Murk and Oliver Jaffe and Olivia Watkins and Olivier Godement and Owen Campbell-Moore and Patrick Chao and Paul McMillan and Pavel Belov and Peng Su and Peter Bak and Peter Bakkum and Peter Deng and Peter Dolan and Peter Hoeschele and Peter Welinder and Phil Tillet and Philip Pronin and Philippe Tillet and Prafulla Dhariwal and Qiming Yuan and Rachel Dias and Rachel Lim and Rahul Arora and Rajan Troll and Randall Lin and Rapha Gontijo Lopes and Raul Puri and Reah Miyara and Reimar Leike and Renaud Gaubert and Reza Zamani and Ricky Wang and Rob Donnelly and Rob Honsby and Rocky Smith and Rohan Sahai and Rohit Ramchandani and Romain Huet and Rory Carmichael and Rowan Zellers and Roy Chen and Ruby Chen and Ruslan Nigmatullin and Ryan Cheu and Saachi Jain and Sam Altman and Sam Schoenholz and Sam Toizer and Samuel Miserendino and Sandhini Agarwal and Sara Culver and Scott Ethersmith and Scott Gray and Sean Grove and Sean Metzger and Shamez Hermani and Shantanu Jain and Shengjia Zhao and Sherwin Wu and Shino Jomoto and Shirong Wu and Shuaiqi and Xia and Sonia Phene and Spencer Papay and Srinivas Narayanan and Steve Coffey and Steve Lee and Stewart Hall and Suchir Balaji and Tal Broda and Tal Stramer and Tao Xu and Tarun Gogineni and Taya Christianson and Ted Sanders and Tejal Patwardhan and Thomas Cunninghman and Thomas Degry and Thomas Dimson and Thomas Raoux and Thomas Shadwell and Tianhao Zheng and Todd Underwood and Todor Markov and Toki Sherbakov and Tom Rubin and Tom Stasi and Tomer Kaftan and Tristan Heywood and Troy Peterson and Tyce Walters and Tyna Eloundou and Valerie Qi and Veit Moeller and Vinnie Monaco and Vishal Kuo and Vlad Fomenko and Wayne Chang and Weiyi Zheng and Wenda Zhou and Wesam Manassra and Will Sheu and Wojciech Zaremba and Yash Patil and Yilei Qian and Yongjik Kim and Youlong Cheng and Yu Zhang and Yuchen He and Yuchen Zhang and Yujia Jin and Yunxing Dai and Yury Malkov},
      year={2024},
      eprint={2410.21276},
      archivePrefix={arXiv},
      primaryClass={cs.CL},
      url={https://arxiv.org/abs/2410.21276}, 
}

@misc{grattafiori2024llama,
    title={The Llama 3 Herd of Models},
    author={Aaron Grattafiori and Abhimanyu Dubey and Abhinav Jauhri and Abhinav Pandey and Abhishek Kadian and Ahmad Al-Dahle and Aiesha Letman and Akhil Mathur and Alan Schelten and Alex Vaughan and Amy Yang and Angela Fan and Anirudh Goyal and Anthony Hartshorn and Aobo Yang and Archi Mitra and Archie Sravankumar and Artem Korenev and Arthur Hinsvark and Arun Rao and Aston Zhang and Aurelien Rodriguez and Austen Gregerson and Ava Spataru and Baptiste Roziere and Bethany Biron and Binh Tang and Bobbie Chern and Charlotte Caucheteux and Chaya Nayak and Chloe Bi and Chris Marra and Chris McConnell and Christian Keller and Christophe Touret and Chunyang Wu and Corinne Wong and Cristian Canton Ferrer and Cyrus Nikolaidis and Damien Allonsius and Daniel Song and Danielle Pintz and Danny Livshits and Danny Wyatt and David Esiobu and Dhruv Choudhary and Dhruv Mahajan and Diego Garcia-Olano and Diego Perino and Dieuwke Hupkes and Egor Lakomkin and Ehab AlBadawy and Elina Lobanova and Emily Dinan and Eric Michael Smith and Filip Radenovic and Francisco Guzmán and Frank Zhang and Gabriel Synnaeve and Gabrielle Lee and Georgia Lewis Anderson and Govind Thattai and Graeme Nail and Gregoire Mialon and Guan Pang and Guillem Cucurell and Hailey Nguyen and Hannah Korevaar and Hu Xu and Hugo Touvron and Iliyan Zarov and Imanol Arrieta Ibarra and Isabel Kloumann and Ishan Misra and Ivan Evtimov and Jack Zhang and Jade Copet and Jaewon Lee and Jan Geffert and Jana Vranes and Jason Park and Jay Mahadeokar and Jeet Shah and Jelmer van der Linde and Jennifer Billock and Jenny Hong and Jenya Lee and Jeremy Fu and Jianfeng Chi and Jianyu Huang and Jiawen Liu and Jie Wang and Jiecao Yu and Joanna Bitton and Joe Spisak and Jongsoo Park and Joseph Rocca and Joshua Johnstun and Joshua Saxe and Junteng Jia and Kalyan Vasuden Alwala and Karthik Prasad and Kartikeya Upasani and Kate Plawiak and Ke Li and Kenneth Heafield and Kevin Stone and Khalid El-Arini and Krithika Iyer and Kshitiz Malik and Kuenley Chiu and Kunal Bhalla and Kushal Lakhotia and Lauren Rantala-Yeary and Laurens van der Maaten and Lawrence Chen and Liang Tan and Liz Jenkins and Louis Martin and Lovish Madaan and Lubo Malo and Lukas Blecher and Lukas Landzaat and Luke de Oliveira and Madeline Muzzi and Mahesh Pasupuleti and Mannat Singh and Manohar Paluri and Marcin Kardas and Maria Tsimpoukelli and Mathew Oldham and Mathieu Rita and Maya Pavlova and Melanie Kambadur and Mike Lewis and Min Si and Mitesh Kumar Singh and Mona Hassan and Naman Goyal and Narjes Torabi and Nikolay Bashlykov and Nikolay Bogoychev and Niladri Chatterji and Ning Zhang and Olivier Duchenne and Onur Çelebi and Patrick Alrassy and Pengchuan Zhang and Pengwei Li and Petar Vasic and Peter Weng and Prajjwal Bhargava and Pratik Dubal and Praveen Krishnan and Punit Singh Koura and Puxin Xu and Qing He and Qingxiao Dong and Ragavan Srinivasan and Raj Ganapathy and Ramon Calderer and Ricardo Silveira Cabral and Robert Stojnic and Roberta Raileanu and Rohan Maheswari and Rohit Girdhar and Rohit Patel and Romain Sauvestre and Ronnie Polidoro and Roshan Sumbaly and Ross Taylor and Ruan Silva and Rui Hou and Rui Wang and Saghar Hosseini and Sahana Chennabasappa and Sanjay Singh and Sean Bell and Seohyun Sonia Kim and Sergey Edunov and Shaoliang Nie and Sharan Narang and Sharath Raparthy and Sheng Shen and Shengye Wan and Shruti Bhosale and Shun Zhang and Simon Vandenhende and Soumya Batra and Spencer Whitman and Sten Sootla and Stephane Collot and Suchin Gururangan and Sydney Borodinsky and Tamar Herman and Tara Fowler and Tarek Sheasha and Thomas Georgiou and Thomas Scialom and Tobias Speckbacher and Todor Mihaylov and Tong Xiao and Ujjwal Karn and Vedanuj Goswami and Vibhor Gupta and Vignesh Ramanathan and Viktor Kerkez and Vincent Gonguet and Virginie Do and Vish Vogeti and Vítor Albiero and Vladan Petrovic and Weiwei Chu and Wenhan Xiong and Wenyin Fu and Whitney Meers and Xavier Martinet and Xiaodong Wang and Xiaofang Wang and Xiaoqing Ellen Tan and Xide Xia and Xinfeng Xie and Xuchao Jia and Xuewei Wang and Yaelle Goldschlag and Yashesh Gaur and Yasmine Babaei and Yi Wen and Yiwen Song and Yuchen Zhang and Yue Li and Yuning Mao and Zacharie Delpierre Coudert and Zheng Yan and Zhengxing Chen and Zoe Papakipos and Aaditya Singh and Aayushi Srivastava and Abha Jain and Adam Kelsey and Adam Shajnfeld and Adithya Gangidi and Adolfo Victoria and Ahuva Goldstand and Ajay Menon and Ajay Sharma and Alex Boesenberg and Alexei Baevski and Allie Feinstein and Amanda Kallet and Amit Sangani and Amos Teo and Anam Yunus and Andrei Lupu and Andres Alvarado and Andrew Caples and Andrew Gu and Andrew Ho and Andrew Poulton and Andrew Ryan and Ankit Ramchandani and Annie Dong and Annie Franco and Anuj Goyal and Aparajita Saraf and Arkabandhu Chowdhury and Ashley Gabriel and Ashwin Bharambe and Assaf Eisenman and Azadeh Yazdan and Beau James and Ben Maurer and Benjamin Leonhardi and Bernie Huang and Beth Loyd and Beto De Paola and Bhargavi Paranjape and Bing Liu and Bo Wu and Boyu Ni and Braden Hancock and Bram Wasti and Brandon Spence and Brani Stojkovic and Brian Gamido and Britt Montalvo and Carl Parker and Carly Burton and Catalina Mejia and Ce Liu and Changhan Wang and Changkyu Kim and Chao Zhou and Chester Hu and Ching-Hsiang Chu and Chris Cai and Chris Tindal and Christoph Feichtenhofer and Cynthia Gao and Damon Civin and Dana Beaty and Daniel Kreymer and Daniel Li and David Adkins and David Xu and Davide Testuggine and Delia David and Devi Parikh and Diana Liskovich and Didem Foss and Dingkang Wang and Duc Le and Dustin Holland and Edward Dowling and Eissa Jamil and Elaine Montgomery and Eleonora Presani and Emily Hahn and Emily Wood and Eric-Tuan Le and Erik Brinkman and Esteban Arcaute and Evan Dunbar and Evan Smothers and Fei Sun and Felix Kreuk and Feng Tian and Filippos Kokkinos and Firat Ozgenel and Francesco Caggioni and Frank Kanayet and Frank Seide and Gabriela Medina Florez and Gabriella Schwarz and Gada Badeer and Georgia Swee and Gil Halpern and Grant Herman and Grigory Sizov and Guangyi and Zhang and Guna Lakshminarayanan and Hakan Inan and Hamid Shojanazeri and Han Zou and Hannah Wang and Hanwen Zha and Haroun Habeeb and Harrison Rudolph and Helen Suk and Henry Aspegren and Hunter Goldman and Hongyuan Zhan and Ibrahim Damlaj and Igor Molybog and Igor Tufanov and Ilias Leontiadis and Irina-Elena Veliche and Itai Gat and Jake Weissman and James Geboski and James Kohli and Janice Lam and Japhet Asher and Jean-Baptiste Gaya and Jeff Marcus and Jeff Tang and Jennifer Chan and Jenny Zhen and Jeremy Reizenstein and Jeremy Teboul and Jessica Zhong and Jian Jin and Jingyi Yang and Joe Cummings and Jon Carvill and Jon Shepard and Jonathan McPhie and Jonathan Torres and Josh Ginsburg and Junjie Wang and Kai Wu and Kam Hou U and Karan Saxena and Kartikay Khandelwal and Katayoun Zand and Kathy Matosich and Kaushik Veeraraghavan and Kelly Michelena and Keqian Li and Kiran Jagadeesh and Kun Huang and Kunal Chawla and Kyle Huang and Lailin Chen and Lakshya Garg and Lavender A and Leandro Silva and Lee Bell and Lei Zhang and Liangpeng Guo and Licheng Yu and Liron Moshkovich and Luca Wehrstedt and Madian Khabsa and Manav Avalani and Manish Bhatt and Martynas Mankus and Matan Hasson and Matthew Lennie and Matthias Reso and Maxim Groshev and Maxim Naumov and Maya Lathi and Meghan Keneally and Miao Liu and Michael L. Seltzer and Michal Valko and Michelle Restrepo and Mihir Patel and Mik Vyatskov and Mikayel Samvelyan and Mike Clark and Mike Macey and Mike Wang and Miquel Jubert Hermoso and Mo Metanat and Mohammad Rastegari and Munish Bansal and Nandhini Santhanam and Natascha Parks and Natasha White and Navyata Bawa and Nayan Singhal and Nick Egebo and Nicolas Usunier and Nikhil Mehta and Nikolay Pavlovich Laptev and Ning Dong and Norman Cheng and Oleg Chernoguz and Olivia Hart and Omkar Salpekar and Ozlem Kalinli and Parkin Kent and Parth Parekh and Paul Saab and Pavan Balaji and Pedro Rittner and Philip Bontrager and Pierre Roux and Piotr Dollar and Polina Zvyagina and Prashant Ratanchandani and Pritish Yuvraj and Qian Liang and Rachad Alao and Rachel Rodriguez and Rafi Ayub and Raghotham Murthy and Raghu Nayani and Rahul Mitra and Rangaprabhu Parthasarathy and Raymond Li and Rebekkah Hogan and Robin Battey and Rocky Wang and Russ Howes and Ruty Rinott and Sachin Mehta and Sachin Siby and Sai Jayesh Bondu and Samyak Datta and Sara Chugh and Sara Hunt and Sargun Dhillon and Sasha Sidorov and Satadru Pan and Saurabh Mahajan and Saurabh Verma and Seiji Yamamoto and Sharadh Ramaswamy and Shaun Lindsay and Shaun Lindsay and Sheng Feng and Shenghao Lin and Shengxin Cindy Zha and Shishir Patil and Shiva Shankar and Shuqiang Zhang and Shuqiang Zhang and Sinong Wang and Sneha Agarwal and Soji Sajuyigbe and Soumith Chintala and Stephanie Max and Stephen Chen and Steve Kehoe and Steve Satterfield and Sudarshan Govindaprasad and Sumit Gupta and Summer Deng and Sungmin Cho and Sunny Virk and Suraj Subramanian and Sy Choudhury and Sydney Goldman and Tal Remez and Tamar Glaser and Tamara Best and Thilo Koehler and Thomas Robinson and Tianhe Li and Tianjun Zhang and Tim Matthews and Timothy Chou and Tzook Shaked and Varun Vontimitta and Victoria Ajayi and Victoria Montanez and Vijai Mohan and Vinay Satish Kumar and Vishal Mangla and Vlad Ionescu and Vlad Poenaru and Vlad Tiberiu Mihailescu and Vladimir Ivanov and Wei Li and Wenchen Wang and Wenwen Jiang and Wes Bouaziz and Will Constable and Xiaocheng Tang and Xiaojian Wu and Xiaolan Wang and Xilun Wu and Xinbo Gao and Yaniv Kleinman and Yanjun Chen and Ye Hu and Ye Jia and Ye Qi and Yenda Li and Yilin Zhang and Ying Zhang and Yossi Adi and Youngjin Nam and Yu and Wang and Yu Zhao and Yuchen Hao and Yundi Qian and Yunlu Li and Yuzi He and Zach Rait and Zachary DeVito and Zef Rosnbrick and Zhaoduo Wen and Zhenyu Yang and Zhiwei Zhao and Zhiyu Ma},
    year={2024},
    eprint={2407.21783},
    archivePrefix={arXiv},
    primaryClass={cs.AI}
}

@misc{makoviychuk2021isaacgymhighperformance,
      title={Isaac Gym: High Performance GPU-Based Physics Simulation For Robot Learning}, 
      author={Viktor Makoviychuk and Lukasz Wawrzyniak and Yunrong Guo and Michelle Lu and Kier Storey and Miles Macklin and David Hoeller and Nikita Rudin and Arthur Allshire and Ankur Handa and Gavriel State},
      year={2021},
      eprint={2108.10470},
      archivePrefix={arXiv},
      primaryClass={cs.RO},
      url={https://arxiv.org/abs/2108.10470}, 
}

@misc{yang2023unisimneuralclosedloopsensor,
      title={UniSim: A Neural Closed-Loop Sensor Simulator}, 
      author={Ze Yang and Yun Chen and Jingkang Wang and Sivabalan Manivasagam and Wei-Chiu Ma and Anqi Joyce Yang and Raquel Urtasun},
      year={2023},
      eprint={2308.01898},
      archivePrefix={arXiv},
      primaryClass={cs.CV},
      url={https://arxiv.org/abs/2308.01898}, 
}

@misc{tobin2018domainrandomizationgenerativemodels,
      title={Domain Randomization and Generative Models for Robotic Grasping}, 
      author={Joshua Tobin and Lukas Biewald and Rocky Duan and Marcin Andrychowicz and Ankur Handa and Vikash Kumar and Bob McGrew and Jonas Schneider and Peter Welinder and Wojciech Zaremba and Pieter Abbeel},
      year={2018},
      eprint={1710.06425},
      archivePrefix={arXiv},
      primaryClass={cs.RO},
      url={https://arxiv.org/abs/1710.06425}, 
}

@misc{chen2022metalearninglanguagemodelincontext,
      title={Meta-learning via Language Model In-context Tuning}, 
      author={Yanda Chen and Ruiqi Zhong and Sheng Zha and George Karypis and He He},
      year={2022},
      eprint={2110.07814},
      archivePrefix={arXiv},
      primaryClass={cs.CL},
      url={https://arxiv.org/abs/2110.07814}, 
}

@misc{min2022metaicllearninglearncontext,
      title={MetaICL: Learning to Learn In Context}, 
      author={Sewon Min and Mike Lewis and Luke Zettlemoyer and Hannaneh Hajishirzi},
      year={2022},
      eprint={2110.15943},
      archivePrefix={arXiv},
      primaryClass={cs.CL},
      url={https://arxiv.org/abs/2110.15943}, 
}

@misc{yu2024metamathbootstrapmathematicalquestions,
      title={MetaMath: Bootstrap Your Own Mathematical Questions for Large Language Models}, 
      author={Longhui Yu and Weisen Jiang and Han Shi and Jincheng Yu and Zhengying Liu and Yu Zhang and James T. Kwok and Zhenguo Li and Adrian Weller and Weiyang Liu},
      year={2024},
      eprint={2309.12284},
      archivePrefix={arXiv},
      primaryClass={cs.CL},
      url={https://arxiv.org/abs/2309.12284}, 
}

@misc{meister2024benchmarkingdistributionalalignmentlarge,
      title={Benchmarking Distributional Alignment of Large Language Models}, 
      author={Nicole Meister and Carlos Guestrin and Tatsunori Hashimoto},
      year={2024},
      eprint={2411.05403},
      archivePrefix={arXiv},
      primaryClass={cs.CL},
      url={https://arxiv.org/abs/2411.05403}, 
}

@misc{melnyk2024distributionalpreferencealignmentllms,
      title={Distributional Preference Alignment of LLMs via Optimal Transport}, 
      author={Igor Melnyk and Youssef Mroueh and Brian Belgodere and Mattia Rigotti and Apoorva Nitsure and Mikhail Yurochkin and Kristjan Greenewald and Jiri Navratil and Jerret Ross},
      year={2024},
      eprint={2406.05882},
      archivePrefix={arXiv},
      primaryClass={cs.LG},
      url={https://arxiv.org/abs/2406.05882}, 
}

@misc{lee2024testtimealignmenthypothesisreweighting,
      title={Test-Time Alignment via Hypothesis Reweighting}, 
      author={Yoonho Lee and Jonathan Williams and Henrik Marklund and Archit Sharma and Eric Mitchell and Anikait Singh and Chelsea Finn},
      year={2024},
      eprint={2412.08812},
      archivePrefix={arXiv},
      primaryClass={cs.LG},
      url={https://arxiv.org/abs/2412.08812}, 
}

@misc{poddar2024personalizingreinforcementlearninghuman,
      title={Personalizing Reinforcement Learning from Human Feedback with Variational Preference Learning}, 
      author={Sriyash Poddar and Yanming Wan and Hamish Ivison and Abhishek Gupta and Natasha Jaques},
      year={2024},
      eprint={2408.10075},
      archivePrefix={arXiv},
      primaryClass={cs.LG},
      url={https://arxiv.org/abs/2408.10075}, 
}

@misc{zhao2024grouppreferenceoptimizationfewshot,
      title={Group Preference Optimization: Few-Shot Alignment of Large Language Models}, 
      author={Siyan Zhao and John Dang and Aditya Grover},
      year={2024},
      eprint={2310.11523},
      archivePrefix={arXiv},
      primaryClass={cs.LG},
      url={https://arxiv.org/abs/2310.11523}, 
}

@misc{shaikh2024showdonttellaligning,
      title={Show, Don't Tell: Aligning Language Models with Demonstrated Feedback}, 
      author={Omar Shaikh and Michelle Lam and Joey Hejna and Yijia Shao and Michael Bernstein and Diyi Yang},
      year={2024},
      eprint={2406.00888},
      archivePrefix={arXiv},
      primaryClass={cs.CL},
      url={https://arxiv.org/abs/2406.00888}, 
}

@misc{siththaranjan2024distributionalpreferencelearningunderstanding,
      title={Distributional Preference Learning: Understanding and Accounting for Hidden Context in RLHF}, 
      author={Anand Siththaranjan and Cassidy Laidlaw and Dylan Hadfield-Menell},
      year={2024},
      eprint={2312.08358},
      archivePrefix={arXiv},
      primaryClass={cs.LG},
      url={https://arxiv.org/abs/2312.08358}, 
}

@inproceedings{cheng-etal-2023-marked,
    title = "Marked Personas: Using Natural Language Prompts to Measure Stereotypes in Language Models",
    author = "Cheng, Myra  and
      Durmus, Esin  and
      Jurafsky, Dan",
    editor = "Rogers, Anna  and
      Boyd-Graber, Jordan  and
      Okazaki, Naoaki",
    booktitle = "Proceedings of the 61st Annual Meeting of the Association for Computational Linguistics (Volume 1: Long Papers)",
    month = jul,
    year = "2023",
    address = "Toronto, Canada",
    publisher = "Association for Computational Linguistics",
    url = "https://aclanthology.org/2023.acl-long.84/",
    doi = "10.18653/v1/2023.acl-long.84",
    pages = "1504--1532"
}

@misc{bai2022constitutionalaiharmlessnessai,
      title={Constitutional AI: Harmlessness from AI Feedback}, 
      author={Yuntao Bai and Saurav Kadavath and Sandipan Kundu and Amanda Askell and Jackson Kernion and Andy Jones and Anna Chen and Anna Goldie and Azalia Mirhoseini and Cameron McKinnon and Carol Chen and Catherine Olsson and Christopher Olah and Danny Hernandez and Dawn Drain and Deep Ganguli and Dustin Li and Eli Tran-Johnson and Ethan Perez and Jamie Kerr and Jared Mueller and Jeffrey Ladish and Joshua Landau and Kamal Ndousse and Kamile Lukosuite and Liane Lovitt and Michael Sellitto and Nelson Elhage and Nicholas Schiefer and Noemi Mercado and Nova DasSarma and Robert Lasenby and Robin Larson and Sam Ringer and Scott Johnston and Shauna Kravec and Sheer El Showk and Stanislav Fort and Tamera Lanham and Timothy Telleen-Lawton and Tom Conerly and Tom Henighan and Tristan Hume and Samuel R. Bowman and Zac Hatfield-Dodds and Ben Mann and Dario Amodei and Nicholas Joseph and Sam McCandlish and Tom Brown and Jared Kaplan},
      year={2022},
      eprint={2212.08073},
      archivePrefix={arXiv},
      primaryClass={cs.CL},
      url={https://arxiv.org/abs/2212.08073}, 
}

@misc{jang2023personalizedsoupspersonalizedlarge,
      title={Personalized Soups: Personalized Large Language Model Alignment via Post-hoc Parameter Merging}, 
      author={Joel Jang and Seungone Kim and Bill Yuchen Lin and Yizhong Wang and Jack Hessel and Luke Zettlemoyer and Hannaneh Hajishirzi and Yejin Choi and Prithviraj Ammanabrolu},
      year={2023},
      eprint={2310.11564},
      archivePrefix={arXiv},
      primaryClass={cs.CL},
      url={https://arxiv.org/abs/2310.11564}, 
}

@misc{sorensen2024roadmappluralisticalignment,
      title={A Roadmap to Pluralistic Alignment}, 
      author={Taylor Sorensen and Jared Moore and Jillian Fisher and Mitchell Gordon and Niloofar Mireshghallah and Christopher Michael Rytting and Andre Ye and Liwei Jiang and Ximing Lu and Nouha Dziri and Tim Althoff and Yejin Choi},
      year={2024},
      eprint={2402.05070},
      archivePrefix={arXiv},
      primaryClass={cs.AI},
      url={https://arxiv.org/abs/2402.05070}, 
}

@misc{zhang2024generativeverifiersrewardmodeling,
      title={Generative Verifiers: Reward Modeling as Next-Token Prediction}, 
      author={Lunjun Zhang and Arian Hosseini and Hritik Bansal and Mehran Kazemi and Aviral Kumar and Rishabh Agarwal},
      year={2024},
      eprint={2408.15240},
      archivePrefix={arXiv},
      primaryClass={cs.LG},
      url={https://arxiv.org/abs/2408.15240}, 
}

@misc{li2022sequenceparallelismlongsequence,
      title={Sequence Parallelism: Long Sequence Training from System Perspective}, 
      author={Shenggui Li and Fuzhao Xue and Chaitanya Baranwal and Yongbin Li and Yang You},
      year={2022},
      eprint={2105.13120},
      archivePrefix={arXiv},
      primaryClass={cs.LG},
      url={https://arxiv.org/abs/2105.13120}, 
}

@misc{yang2024contextparallelismscalablemilliontoken,
      title={Context Parallelism for Scalable Million-Token Inference}, 
      author={Amy Yang and Jingyi Yang and Aya Ibrahim and Xinfeng Xie and Bangsheng Tang and Grigory Sizov and Jeremy Reizenstein and Jongsoo Park and Jianyu Huang},
      year={2024},
      eprint={2411.01783},
      archivePrefix={arXiv},
      primaryClass={cs.DC},
      url={https://arxiv.org/abs/2411.01783}, 
}

@misc{li2024syntheticdataalmostscratch,
      title={Synthetic Data (Almost) from Scratch: Generalized Instruction Tuning for Language Models}, 
      author={Haoran Li and Qingxiu Dong and Zhengyang Tang and Chaojun Wang and Xingxing Zhang and Haoyang Huang and Shaohan Huang and Xiaolong Huang and Zeqiang Huang and Dongdong Zhang and Yuxian Gu and Xin Cheng and Xun Wang and Si-Qing Chen and Li Dong and Wei Lu and Zhifang Sui and Benyou Wang and Wai Lam and Furu Wei},
      year={2024},
      eprint={2402.13064},
      archivePrefix={arXiv},
      primaryClass={cs.CL},
      url={https://arxiv.org/abs/2402.13064}, 
}

@misc{dubois2024lengthcontrolledalpacaevalsimpleway,
      title={Length-Controlled AlpacaEval: A Simple Way to Debias Automatic Evaluators}, 
      author={Yann Dubois and Balázs Galambosi and Percy Liang and Tatsunori B. Hashimoto},
      year={2024},
      eprint={2404.04475},
      archivePrefix={arXiv},
      primaryClass={cs.LG},
      url={https://arxiv.org/abs/2404.04475}, 
}

@misc{goodreads,
  author       = {{Goodreads}},
  title        = {Goodreads: Book Reviews, Recommendations, and Discussion},
  year         = {2025},
  url          = {https://www.goodreads.com/},
  note         = {Accessed: 2025-02-15}
}

@misc{imdb,
  author       = {{IMDb}},
  title        = {IMDb: Ratings, Reviews, and Where to Watch the Best Movies \& TV Shows},
  year         = {2025},
  url          = {https://www.imdb.com/},
  note         = {Accessed: 2025-02-15}
}

@misc{myanimelist,
  author       = {{MyAnimeList}},
  title        = {MyAnimeList: Track, Discover, and Discuss Anime \& Manga},
  year         = {2025},
  url          = {https://myanimelist.net/},
  note         = {Accessed: 2025-02-15}
}

@misc{qwen2025qwen25technicalreport,
      title={Qwen2.5 Technical Report}, 
      author={Qwen and : and An Yang and Baosong Yang and Beichen Zhang and Binyuan Hui and Bo Zheng and Bowen Yu and Chengyuan Li and Dayiheng Liu and Fei Huang and Haoran Wei and Huan Lin and Jian Yang and Jianhong Tu and Jianwei Zhang and Jianxin Yang and Jiaxi Yang and Jingren Zhou and Junyang Lin and Kai Dang and Keming Lu and Keqin Bao and Kexin Yang and Le Yu and Mei Li and Mingfeng Xue and Pei Zhang and Qin Zhu and Rui Men and Runji Lin and Tianhao Li and Tianyi Tang and Tingyu Xia and Xingzhang Ren and Xuancheng Ren and Yang Fan and Yang Su and Yichang Zhang and Yu Wan and Yuqiong Liu and Zeyu Cui and Zhenru Zhang and Zihan Qiu},
      year={2025},
      eprint={2412.15115},
      archivePrefix={arXiv},
      primaryClass={cs.CL},
      url={https://arxiv.org/abs/2412.15115}, 
}

@misc{zheng2024sglangefficientexecutionstructured,
      title={SGLang: Efficient Execution of Structured Language Model Programs}, 
      author={Lianmin Zheng and Liangsheng Yin and Zhiqiang Xie and Chuyue Sun and Jeff Huang and Cody Hao Yu and Shiyi Cao and Christos Kozyrakis and Ion Stoica and Joseph E. Gonzalez and Clark Barrett and Ying Sheng},
      year={2024},
      eprint={2312.07104},
      archivePrefix={arXiv},
      primaryClass={cs.AI},
      url={https://arxiv.org/abs/2312.07104}, 
}

@misc{kwon2023efficientmemorymanagementlarge,
      title={Efficient Memory Management for Large Language Model Serving with PagedAttention}, 
      author={Woosuk Kwon and Zhuohan Li and Siyuan Zhuang and Ying Sheng and Lianmin Zheng and Cody Hao Yu and Joseph E. Gonzalez and Hao Zhang and Ion Stoica},
      year={2023},
      eprint={2309.06180},
      archivePrefix={arXiv},
      primaryClass={cs.LG},
      url={https://arxiv.org/abs/2309.06180}, 
}

@article{wu2024aligning,
  title={Aligning LLMs with Individual Preferences via Interaction},
  author={Wu, Shujin and Fung, May and Qian, Cheng and Kim, Jeonghwan and Hakkani-Tur, Dilek and Ji, Heng},
  journal={arXiv preprint arXiv:2410.03642},
  year={2024}
}

@article{yang2024rewards,
  title={Rewards-in-Context: Multi-objective Alignment of Foundation Models with Dynamic Preference Adjustment},
  author={Yang, Rui and Pan, Xiaoman and Luo, Feng and Qiu, Shuang and Zhong, Han and Yu, Dong and Chen, Jianshu},
  journal={International Conference on Machine Learning},
  year={2024}
}
\clearpage
\appendix
\section{Appendix}
\subsection{Hyperparameters for \methodname{}}

\begin{table*}[h]
    \centering
    \renewcommand{\arraystretch}{1.2} 
    \setlength{\tabcolsep}{8pt} 
    \begin{tabular}{l c}
        \toprule
        \textbf{Name} & \textbf{Value} \\
        \midrule
        Learning Rate (SFT/Pref-FT) & \(1e^{-5}, 1e^{-6}, \mathbf{1e^{-7}}\) \\
        Learning Rate (IPO) & \(1e^{-5}, \mathbf{1e^{-6}}, 1e^{-7}\) \\
        Beta (IPO) & \(0.1, 0.05, 0.01, \mathbf{0.005}, 0.001\) \\
        Number of Shots & \(4, \mathbf{8}\) \\
        Model Name & Llama 3.2 3B Instruct~\citep{grattafiori2024llama} \\
        \bottomrule
    \end{tabular}
    \caption{Sweep over hyperparameters for \methodname, recommended hyperparameters in bold.}
    \label{tab:hyperparameters}
\end{table*}

\subsubsection{Additional Ablations}
We perform two ablations to study the impact of the size of the preference dataset and number of few-shot examples on performance. We see a monotonic increase in performance over the size and the number of fewshot examples in the Roleplay dataset.
\begin{table}[ht]
    \centering
    \begin{subtable}[t]{0.45\textwidth}
        \centering
        \begin{tabular}{c|c}
            \hline
            \textbf{Preference Data (\%)} & \textbf{Winrate (\%)} \\
            \hline
            10  & 70.1 \\
            25  & 69.5 \\
            50  & 78.3 \\
            100 & 82.6 (reported) \\
            \hline
        \end{tabular}
        \caption{Varying percentage of preference data.}
        \label{tab:preference_data_ablation}
    \end{subtable}
    \hspace{0.05\textwidth}
    \begin{subtable}[t]{0.45\textwidth}
        \centering
        \begin{tabular}{c|c}
            \hline
            \textbf{Few-Shot Examples} & \textbf{Winrate (\%)} \\
            \hline
            1 & 65.7 \\
            2 & 69.3 \\
            4 & 72.1 \\
            8 & 82.6 (reported) \\
            \hline
        \end{tabular}
        \caption{Varying number of few-shot examples.}
        \label{tab:few_shot_ablation}
    \end{subtable}
    \caption{Ablation studies on roleplay task winrates with held-out synthetic users: (a) effect of preference data percentage per user, and (b) effect of number of few-shot examples.}
    \label{tab:ablation_combined}
\end{table}
\subsection{Sample Personalized Responses}
\label{sec:app_samples}
\begin{figure*}
    \centering
    \includegraphics[width=0.8\linewidth]{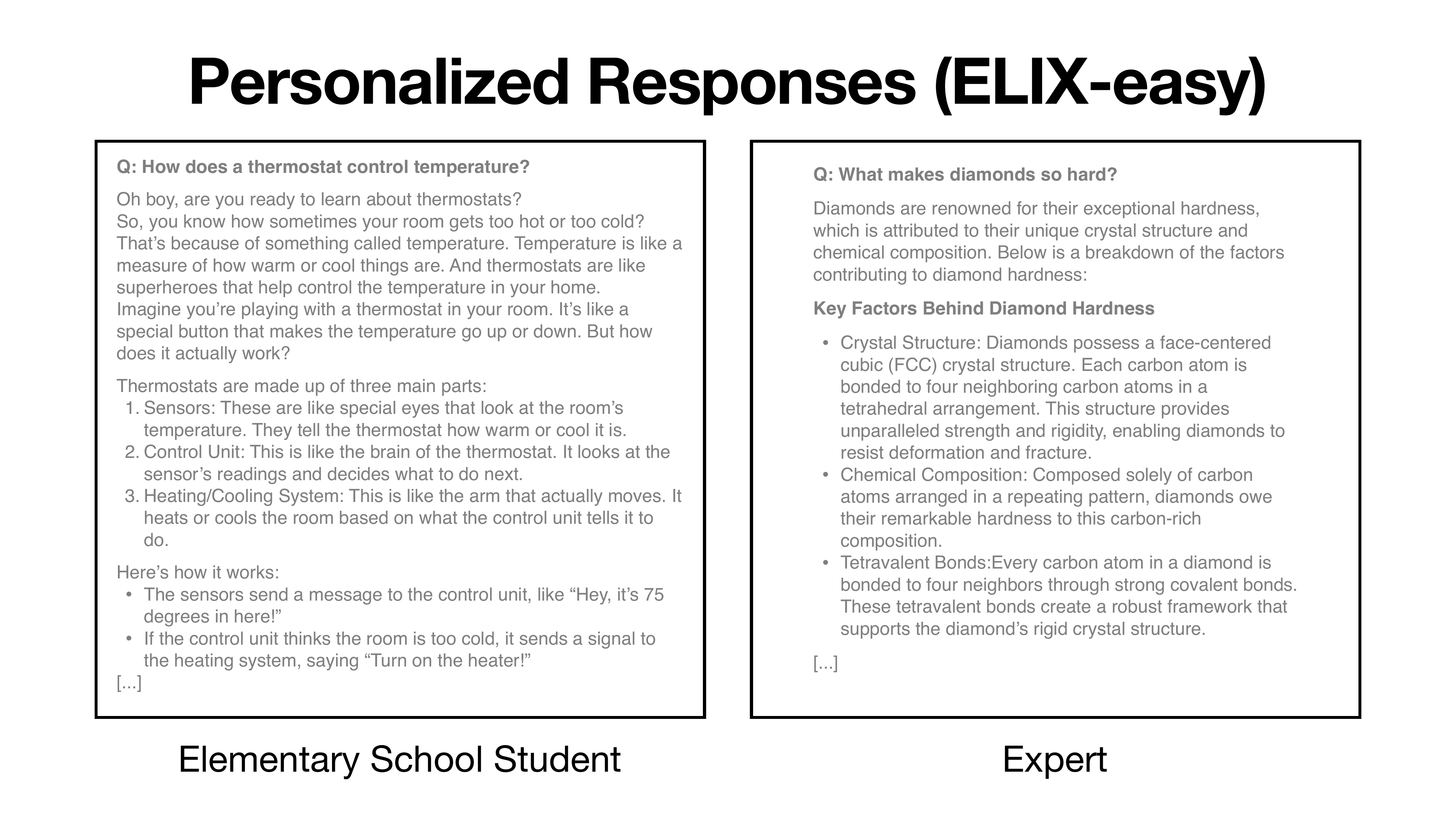}
    \includegraphics[width=0.8\linewidth]{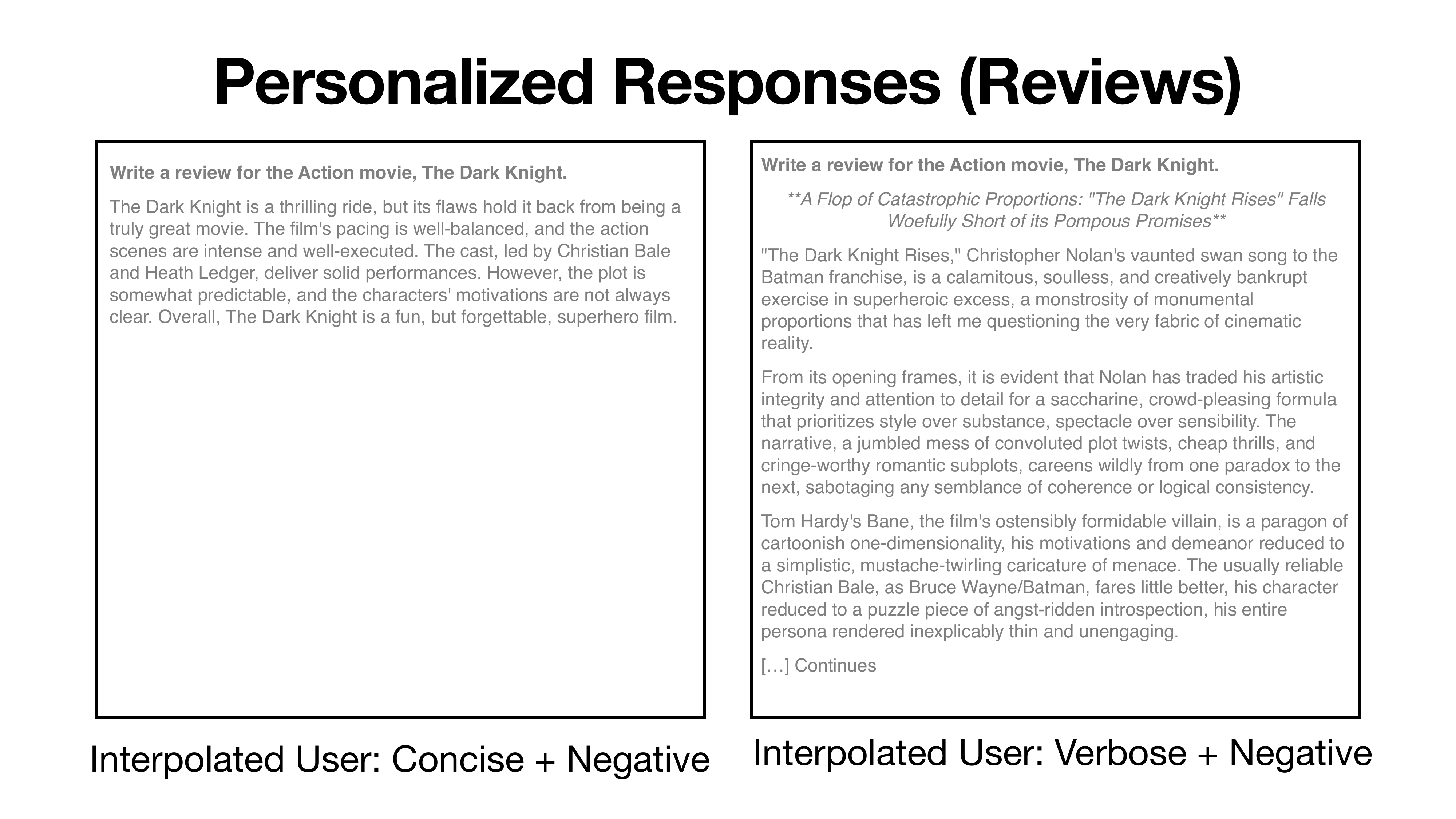}
    \caption{\footnotesize\textbf{Sample Personalized Response for ELIX (top) and Reviews (bottom).}}
    \label{fig:personalized_responses}
\end{figure*}

We provide sample responses from \methodname\ in Figure~\ref{fig:personalized_responses} across the 3 tasks that were studied (ELIX, Reviews, and Roleplay). We additionally include the oracle scoring description for each response, demonstrating \methodname's ability to personalize. Additionally, for the Review task, we show responses for interpolated users which were not found in the preference dataset, showing the ability of \methodname\ to generalize and be controllable.

\subsubsection{Roleplay User Description Rationalization (RAT)}
\label{sec:user_desc_rat}
\begin{lstlisting}[frame=single,breaklines=true]
QUESTION:
What are some good volunteer opportunities for me?

TRUE PERSONA:
A 53-year-old woman living in North America, an avid traveler and writer who shares her adventures through a popular travel blog. She loves spontaneous adventures and values long-term financial security. She is at a point in her career where opportunities for advancement might be time-sensitive. She enjoys movies that challenge perspectives and spark discussions. She prioritizes well-being over being fully informed. She is interested in the van life movement. She appreciates creativity and variety in meal options. She enjoys engaging in creative and tactile activities and prefers traditional learning methods. She values compassionate and responsible choices. She enjoys exploring new and dynamic experiences. She is concerned about fast food. She values mental well-being and mindfulness. She enjoys attending events and conferences to meet new people. She appreciates practicality and health. She enjoys shows with dark humor and suspenseful storytelling.

RAT GENERATED PERSONA:
Based on the examples provided, the user appears to value a more holistic and experiential approach to learning and personal growth. They seem to prioritize human connection, nature, and personal development over technology and material wealth.

This user is likely someone who is interested in personal stories, emotions, and experiences, as evident from their appreciation for Sylvia Plath's poetry and the suggestion to weave storytelling into public speaking. They may also be someone who values authenticity, simplicity, and community, as seen in their suggestion of house swapping and volunteering as alternative vacation options.

In terms of lifestyle, this user likely prefers a balanced and flexible approach to life, allowing for a mix of work, personal growth, and leisure activities. They may also be someone who values environmental sustainability and social responsibility, as hinted at by their consideration of eco-friendly home improvements and sustainable financial planning. Overall, this user seems to be a thoughtful, empathetic, and nature-loving individual who prioritizes human connections, experiences, and personal growth over material possessions and technological advancements.

FINAL RESPONSE:
Based on your affinity for human connections, personal growth, and environmental sustainability, I'd like to suggest the following volunteer opportunities that might resonate with you:

**Community Garden or Urban Farm**: Many organizations in your area offer community gardens or urban farms where you can help cultivate and maintain a shared space, promoting sustainability and food security. This allows you to connect with nature and contribute to your community's well-being.

**Nature Reserves or Conservation Efforts**: Consider volunteering with local wildlife conservation organizations, such as bird watching tours, tree planting initiatives, or habitat restoration projects. This will enable you to work outdoors, connect with others who share your passion for nature, and contribute to preserving the environment.

**Community Centers or Environmental Education Programs**: Many community centers offer various programs, such as gardening workshops, environmental education classes, or sustainable living workshops. By volunteering at these centers, you can share your knowledge, learn from others, and make a positive impact on your community.

**Animal Shelters or Wildlife Rehabilitation Centers**: If you're an animal lover, consider volunteering at a local animal shelter or wildlife rehabilitation center. You'll have the opportunity to care for animals, interact with like-minded individuals, and contribute to animal welfare and conservation efforts.

**Neighborhood Events or Community Festivals**: Look for opportunities to volunteer at local events, such as festivals, farmers' markets, or street fairs. This allows you to engage with your community, promote sustainability, and build connections with neighbors through shared passion for environmental causes.

These opportunities will not only nurture your spirit of volunteering but also help you foster meaningful connections with nature, others, and your community while nurturing a sense of personal growth and fulfillment.

\end{lstlisting}

\vspace{-0.2cm}
\subsection{Additional Details on Dataset Construction and Evaluation}
\vspace{-0.2cm}
\label{sec:app_dataset_eval}
\begin{figure*}
    \centering
    \includegraphics[width=\linewidth]{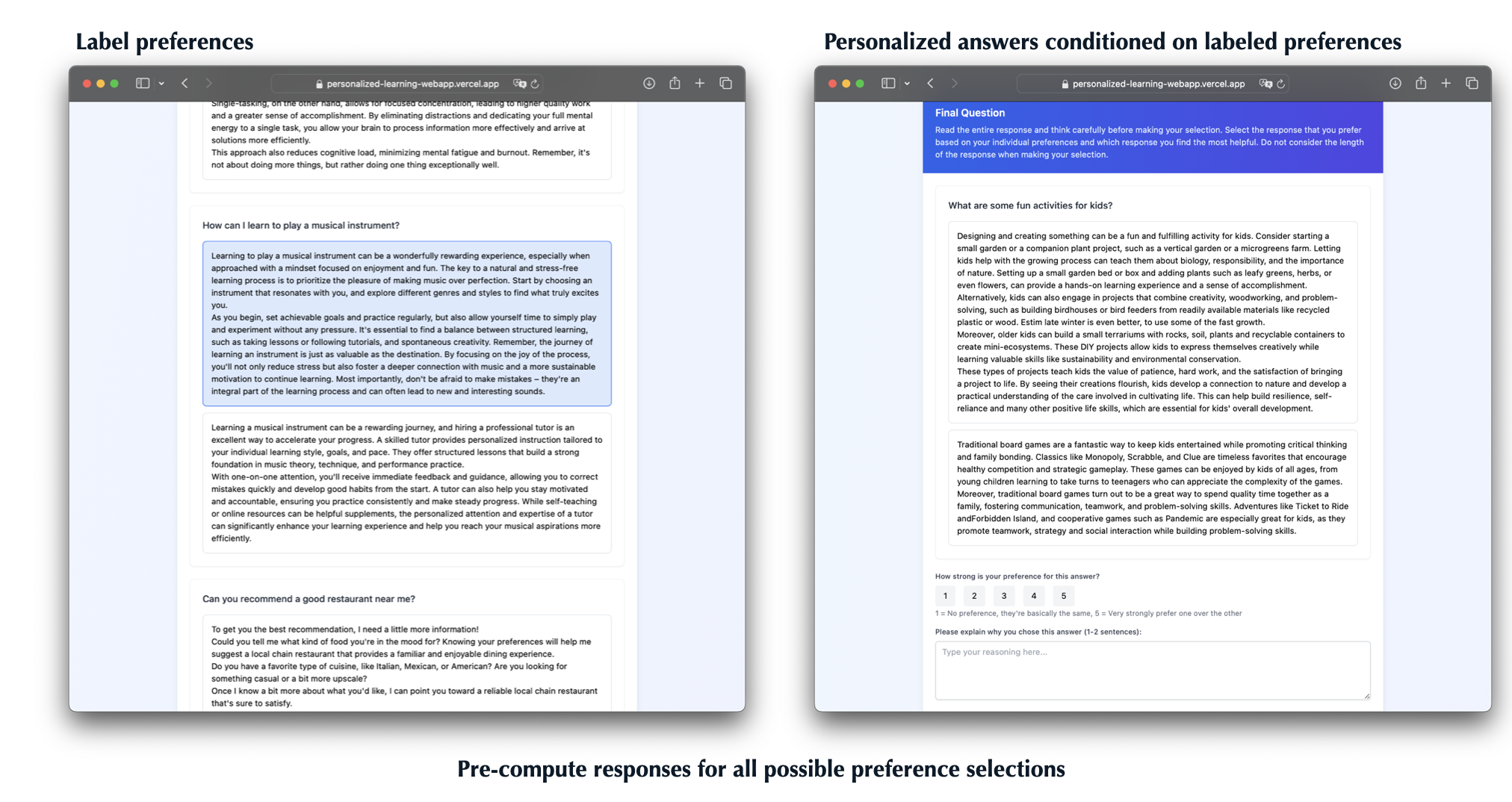}
    \caption{\footnotesize \textbf{An overview of the Human Study Interface.} First, users label a set of preferences. Then, a set of personalized answers are provided, conditioned on label preferences.}
    \vspace{-0.2cm}
    \label{fig:human_app}
    \vspace{-0.3cm}
\end{figure*}

\paragraph{Dataset Details}
We include a representative subset of the roleplay task and a jupyter notebook demonstrating shot construction for training and evaluation in the supplementary material (due to size restrictions). We will release the full datasets for each task in the final release of the paper. Below in Figure~\ref{fig:samp_pref}, we provide a qualitative example of a preference from the larger persona dataset, with details about the scoring persona and the viewpoints that underlie the responses. Additionally in Figure~\ref{fig:samp_personas} and Figure~\ref{fig:samp_viewpoint}, we showcase the qualitative diversity and alignment of viewpoints and personas present in the Roleplay task, enabling for preferences to have both diversity and structure.

\begin{figure*}[htbp]
    \centering
    \includegraphics[width=\linewidth]{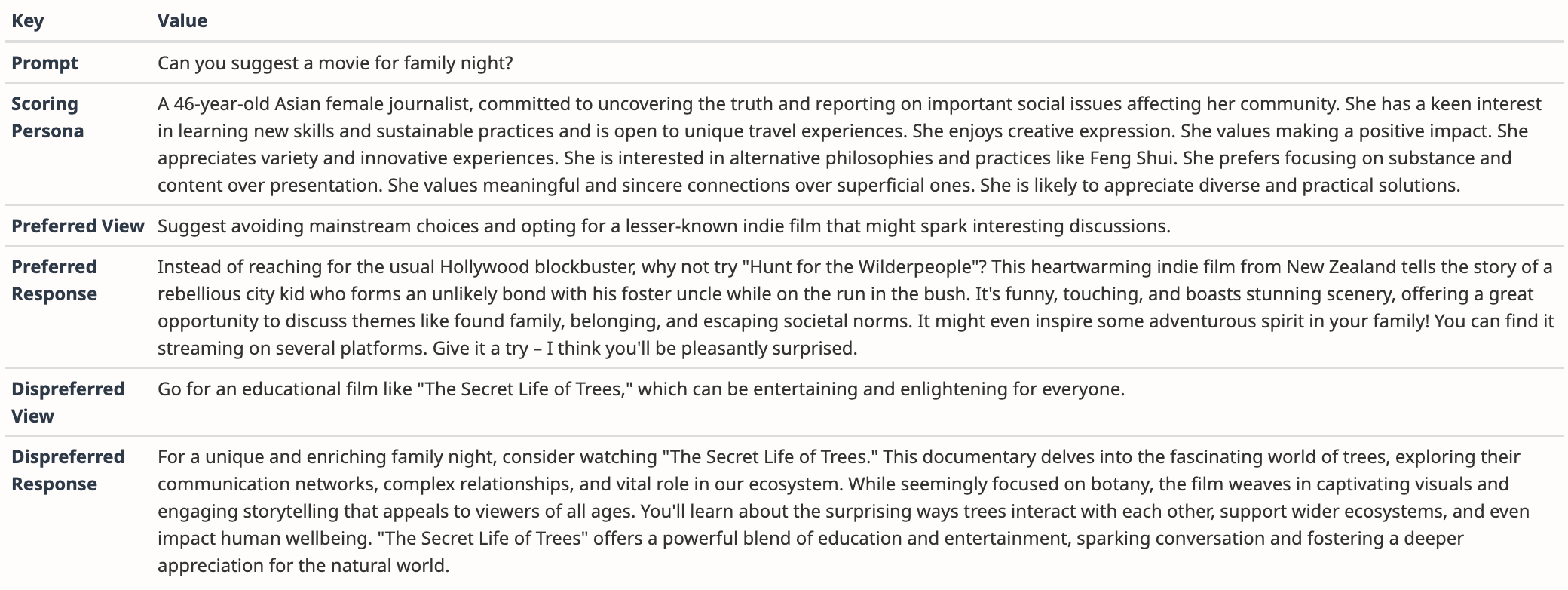}
    \caption{\footnotesize \textbf{Sample Preference (Roleplay).} In this preference example, we provide a sample prompt, scoring persona, (dis)preferred view, and (dis)preferred response.}
    \vspace{-0.2cm}
    \label{fig:samp_pref}
\end{figure*}

\begin{figure*}[htbp]
    \centering
    \includegraphics[width=\linewidth]{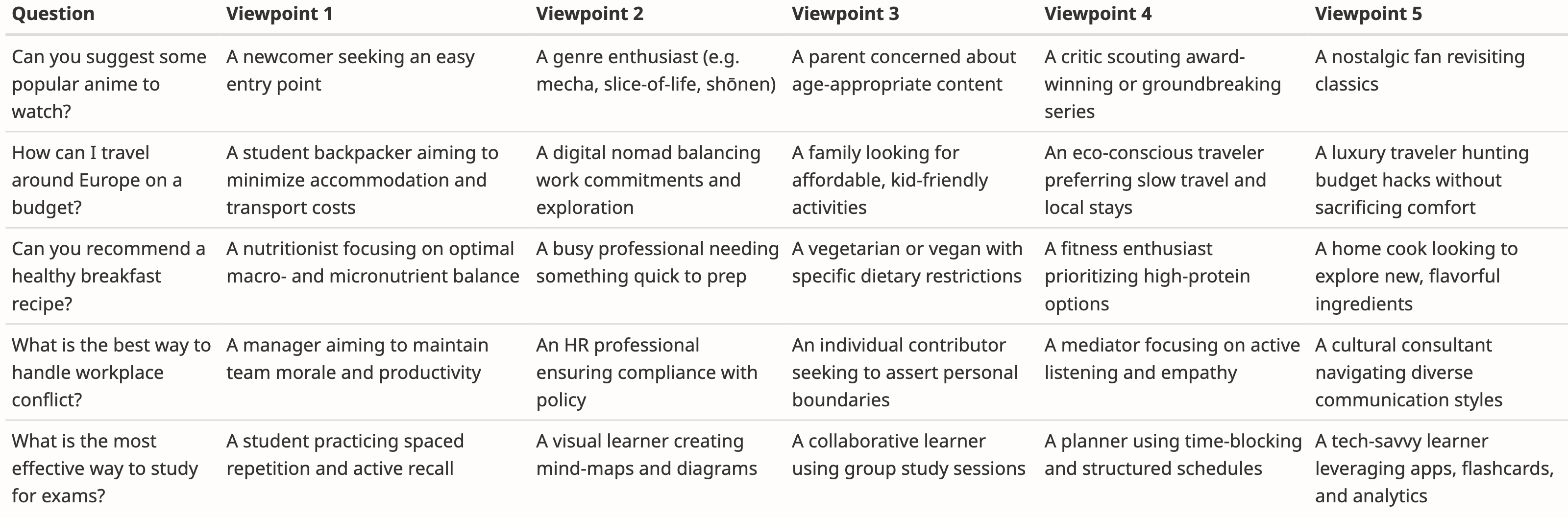}
    \caption{\footnotesize \textbf{Sample Viewpoints (Roleplay).} For a given question, a diversity of viewpoints for a response can be inferred to create a preference dataset that encompasses a wide-range of opinions.}
    \vspace{-0.2cm}
    \label{fig:samp_viewpoint}
\end{figure*}

\begin{figure*}[htbp]
    \centering
    \includegraphics[width=\linewidth]{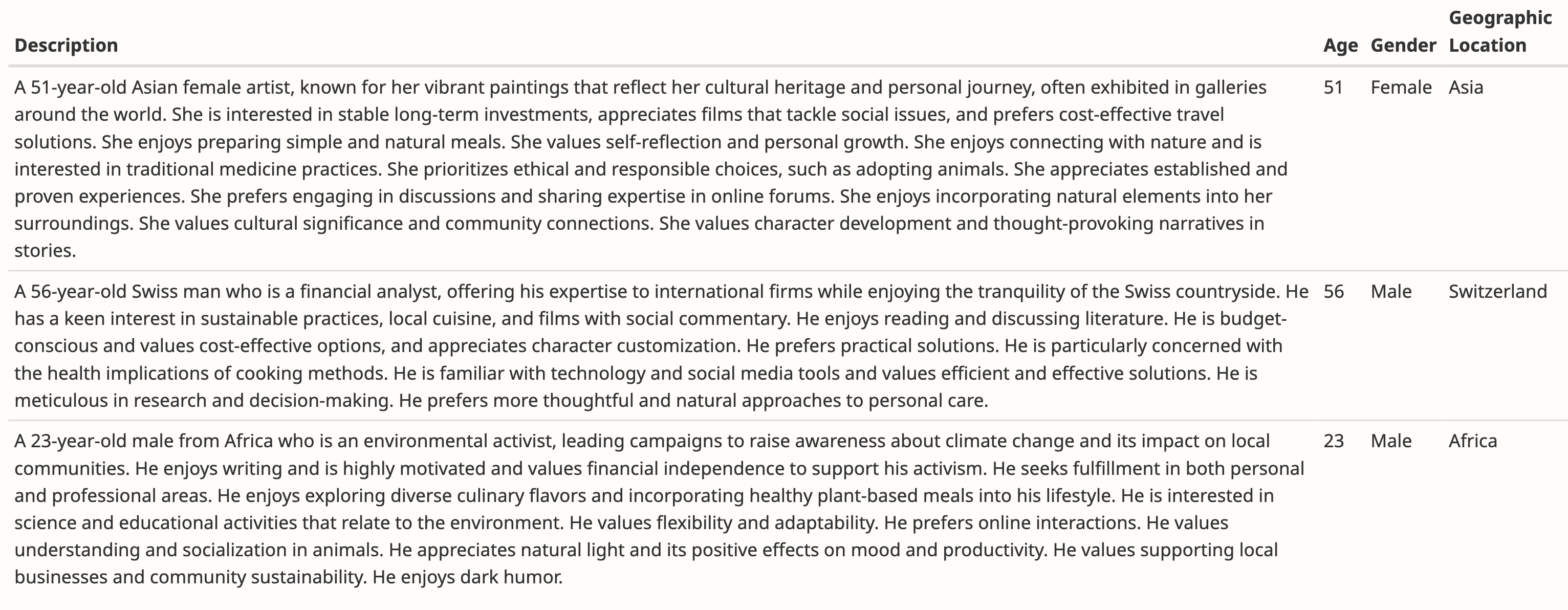}
    \caption{\footnotesize \textbf{Sample Personas (Roleplay).} A comprehensive description of the user is iteratively refined from preference pairs for that user, seeded with attributes of age, gender, and geographic location.}
    \vspace{-0.2cm}
    \label{fig:samp_personas}
\end{figure*}

\paragraph{Evaluation Prompt for Synthetic Preferences}
We used GPT-4o as a Judge using a modified variant of the Alpaca Eval~\citep{dubois2024alpacafarm,dubois2024lengthcontrolledalpacaevalsimpleway} Prompt to be aware of a user description when scoring preference examples.

Here is the system prompt:
\begin{lstlisting}[frame=single,breaklines=true]
You are a highly efficient assistant, who evaluates and selects the best large language model (LLMs) based on the quality of their responses to a given instruction. This process will be used to create a leaderboard reflecting the most accurate and human-preferred answers.
\end{lstlisting}

Here is the user prompt:
\begin{lstlisting}[frame=single,breaklines=true]
You are tasked with evaluating the outputs of multiple large language models to determine which model produces the best response from a human perspective.

## Instructions

You will receive:
1. A **User Instruction**: This is the query or task provided to the models.
2. **Model Outputs**: Unordered responses from different models, each identified by a unique model identifier.
3. A **User Description**: This describes the user's preferences or additional context to guide your evaluation.

Your task is to:
1. Evaluate the outputs based on quality and relevance to the user's instruction and description.
2. Select the best output that meets the user's needs.

## Input Format

### User Instruction
{QUESTION}

### Model Outputs
- Model "m": {RESPONSE_A}
- Model "M": {RESPONSE_B}

### User Description
{USER_DESCRIPTION}

## Task

From the provided outputs, determine which model produces the best response. Output only the model identifier of the best response (either `m` or `M`) with no additional text, quotes, spaces, or new lines.

## Best Model Identifier
\end{lstlisting}

\paragraph{Additional Human Study Details}
As shown in Alpaca Eval 2.0~\citep{dubois2024lengthcontrolledalpacaevalsimpleway}, several biases can affect the evaluation of language models such as length, format, and more. For this reason, we took action to normalize both \methodname\ and baselines in 3 different categories. First, length is an evaluation bias. For this reason, we computed the average length of responses from \methodname\ and prompted the base model during evaluation to keep its responses around the average length in words ($\approx 250$ words). For the SFT baseline, we found that this was consistent with \methodname\ since it was fine-tuned on the same preference dataset. Additionally, due to context length restrictions and the instruction following abilities of smaller open-source LLMs, we decided to have formatting be consistent as paragraphs rather than markdown for the Roleplay task. Thus, we similarly prompted the Base model with this behavior. Finally, a differing number of views can also skew the evaluation, as a large proportion of users seem to prefer direct answers. Additionally, if more views are presented, a user may prefer just one of the many views provided, skewing evaluation. Thus, we ensure that when two responses are compared, they have the same number of views. In future, work, it would be interesting to consider how to relax some of the design decisions needed for the human study. We additionally provide screenshots of the human study interface in Figure~\ref{fig:human_app}.

Below is the full text of instructions given to the participants: \\
"This is a study about personalization. You will be asked to read a set of 20 questions (9 on the first page, 11 on the second page). For each question, there are two responses. Please select the response that you prefer. Make this selection based on your individual preferences and which response you find the most helpful. Read the entire response and think carefully before making your selection."

We utilize the demographic information that Prolific provides for each user such as their age group, continent and gender to chose questions but do not store that information about the user. We collect no identifying information about the user and will not make any of the individual preferences from a user public. We pay each user a fair wage subject to the current region that we reside in. We received consent from the people whose data we are using and curating as the very first question in our survey. The demographic and geographic characteristics of the annotator population is exactly the same as Prolific. We do no filtering of this at all.

\vspace{-0.2cm}
\subsection{Training Details and Hyperparameters for \methodname\ and baselines}
\vspace{-0.2cm}
Similar to DPO~\citep{rafailov2023direct} and IPO~\citep{2023arXiv231012036G}, we trained \methodname\ in a two stage manner. The first stage is Fewshot Pref-FT, increasing the likelihood of the preferred response. The second stage is Fewshot IPO, initialized from the checkpoint of Fewshot Pref-FT. One epoch of the dataset was performed for each stage. For the IPO baseline, we followed a similar procedure. Additional hyperparameters can be found in Table~\ref{tab:hyperparameters}. 

\vspace{-0.2cm}
\subsection{Additional Details of Setup for Reproducability}
\vspace{-0.2cm}
We used both code, models, and data as scientific artifacts. In particular, for code, we built off of the \href{https://github.com/eric-mitchell/direct-preference-optimization}{codebase} from \citet{rafailov2023direct}, with an Apache 2.0 license. We additionally adapted our evaluation script from Alpaca EVAL, including the prompt, and other criterion for evaluation and normalization. We have reported the implementation details for synthetic evaluation in Section~\ref{sec:dom_rand} and human study evaluation in Section ~\ref{sec:app_dataset_eval}.

For models, we used a combination of open-source and closed-source models. The models that we used for sampling data are the Llama family of models~\citep{grattafiori2024llama} (Llama 3.2 3b, Llama 3.1 8b, Llama 3.3 70b) with the llama license (3.1, 3.2, 3.3), the Qwen family of models~\citep{qwen2025qwen25technicalreport} (Qwen 2.5 3b, Qwen 2.5 32b, Qwen 2.5 72b) with the qwen license, the Gemma 2 family of models~\citep{gemmateam2024gemma2improvingopen} (Gemma 2 2b, Gemma 2 9b, and Gemma 2 27b) with the gemma license, and the OpenAI~\citep{openai2024gpt4ocard} family of models (GPT4o, GPT4o-mini) with the OpenAI API License (based off of the MIT License). We used SGLang~\citep{zheng2024sglangefficientexecutionstructured} and VLLM~\citep{kwon2023efficientmemorymanagementlarge} for model inference. For training, we used 1 node of A100 GPUs (8 GPUs) for 8 hours for each experiment with FSDP. Cumulatively, we used approximately 4000 hours of GPU hours for ablations over dataset, architecture design and other details. 

With respect to the dataset, for questions for the review dataset, we sourced media names from IMDb~\citep{imdb}, Goodreads~\citep{goodreads}, and MyAnimeList~\citep{myanimelist}. We define the domains in more detail in section~\ref{sec:domains}. Seed questions for ELIX were human generated, sourced from Prolific. The dataset is entirely in English, with some artifacts of Chinese from the Qwen model family, which will be filtered out for the final release of the dataset. None of this data has identifying information about individual people or offensive content as the dataset was sourced from instruction and safety-tuned models, with each step of the dataset having a manual check of the inputs and outputs. In terms of statistics of the dataset, the review dataset has 130K train/dev examples and 32.4K test examples, the ELIX-easy dataset has 235K train/dev examples and 26.1K test examples, the ELIX-hard dataset has 267K train/dev examples and 267K test examples, and the roleplay dataset has 362K train/dev examples and 58.2K test examples, with a total of 1.378 million examples. For our statistics, we reported the average winrate \% for each method on both synthetic and human evals, following prior work in alignment like AlpacaFarm~\citep{dubois2024alpacafarm}. 

Each of the artifacts above was consistent with its intended use and the code, models, and datasets should be usable outside of research contexts.

\vspace{-0.2cm}
\subsection{Synthetic Data is NOT limited by what is internalized by the LLM}
\vspace{-0.2cm}
Though the seed persona is instantiated and refined with an LLM, one part of the refinement strategy that potentially mitigates the stereotype concern that you have raised is that we randomly select a response to be preferred from a choice of two viewpoint-conditioned responses to augment the seed persona. Therefore, through the refinement process, we recover a persona description that could map to any permutation of the $2^N$ preferences, allowing for more expressivity than what is internalized by the LLM. Additionally, in the viewpoint generation process, we ask the model to list multiple viewpoints for a particular question, which allows the model to elicit a diverse set of possible responses to score and iteratively refine the persona with. This additionally reduces the occurrence of “stereotypical personas”, allowing for more nuanced answers for a particular question. In Figure~\ref{fig:samp_viewpoint}, we list 3 sample personas to qualitatively show their diverse nature.

\vspace{-0.2cm}
\subsection{Sampling of Preferences per User}
\vspace{-0.2cm}
As seen in Algorithm~\ref{alg:fspo}, line 5, for each user, we sample a subset of the user's preferences to construct the few-shot preferences for that user during training. During training, we revisit the user and resample a new subset of preferences. In Table 7, we show an ablation over the number of few-shot preferences that are sampled, and do see gains with the number of preferences conditioned on. For our synthetic evaluation, we match this form of sampling, drawing multiple sets of few-shot preferences per held-out user and averaging over the set to construct the win-rate per user, which we further aggregate over all users. For the human study, due to cost constraints, we ask participants to label a fixed set of preference pairs in the first stage of our study, used as the few-shot examples. Then, for several held-out questions, we evaluate for this fixed set a response from FSPO and a baseline model.

This training and evaluation procedure mitigates the concern that the choice of the N few shot examples impacts performance.

\vspace{-0.2cm}
\subsection{Addressing the Additional Overhead of Few-Shot Personalization}
\vspace{-0.2cm}
Few-shot preferences do expand the context requirements of an LLM. One approach to mitigate this is the RAT prediction, which can be inferred from the user’s preferences and may be shorter than the preferences themselves to condition on. Furthermore, this can be cached for a user to mitigate latency issues and used across different prompts. Finally, models today are continuing to scale the length of their context (such as Gemini 2.5 pro having over a 1 million tokens in context) so this may be a small price to pay with respect to the overall context.

\vspace{-0.2cm}
\subsection{Limitations}
\vspace{-0.2cm}
Our human study was preliminary with control over the questions that a user may ask, format normalization where formatting details such as markdown are removed, and view normalization comparing the same number of viewpoints for both \methodname\ and the baselines. To the best of our knowledge, we are the first to perform such a human study for personalization to open-ended question answering. Future work should do further ablations with human evaluation for personalization. Additionally, due to compute constraints, we work with models in the parameter range of 3B (specifically Llama 3.2 Instruct 3B) with a limited context window of 128K, and without context optimization such as sequence parallelism~\citep{li2022sequenceparallelismlongsequence, yang2024contextparallelismscalablemilliontoken}, further limiting the effective context window. It is an open question on how fine-tuning base models with better long-context and reasoning capabilities would help with \methodname\ for personalization, such as the 2M context window of Gemini  Flash Thinking models, especially in the case of RAT. 

\vspace{-0.2cm}
\subsection{\methodname{} on Human Preference Dataset (Prism Alignment)}
\label{sec:prism_study}
\vspace{-0.2cm}
We have run FSPO on the PRISM Alignment Dataset. For evaluation, we evaluate FSPO as a reward model (leveraging the duality of DPO and IPO) by comparing the log likelihood of the preferred response and dispreferred response on held-out preferences. On this dataset, we achieve a reward prediction accuracy of 82.8\%, whereas population based approaches such as IPO achieve a reward prediction accuracy of 61.7\%, showcasing the efficacy of the method in generalizing to a held-out user. There is no protocol for evaluating generated responses on PRISM, as the survey provided per user is highly underspecified, providing little to no details about the user for response evaluation.

\vspace{-0.2cm}
\subsection{Additional Discussion of the Use of Synthetic Data}
\vspace{-0.2cm}
\label{sec:synthetic_claim}

It would be ideal to use a large-scale real user preference dataset suitable for developing and testing robust personalization systems. Unfortunately, in the open-source community, no such high-quality dataset exists, necessitating the generation of a synthetic preference dataset. In the related work, we do consider a prior human collected dataset, the Prism Alignment Dataset [3], where we find that a proportion of the prompts are of lower quality (such as including conspiracies such as “i think the moon landing was faked”) and quite distinct from questions that a user would ask an assistant, focusing on value-based personalization (such as “Who is right in the Hamas-Israeli war? Hamas or the Israelis?”), which have troubling ethical considerations. In contrast, the Roleplay synthetic dataset studies more natural, recommendation style questions that involve personalization such as “What should I do this weekend in San Francisco?” or “Can you recommend a good podcast?”, synthetically augmented from seed human generated questions.

Thus, the synthetic data pipeline from FSPO can be a practical solution for scenarios where high-quality, task-specific preference data is unavailable, sparse, or lacks diversity. In these situations, our approach can supplement and augment existing real data, rather than merely replacing it. Below, we will describe some real-world problem instances where FSPO can be beneficial.

\textbf{1. Cold Start Problem} One advantage of the synthetic construction proposed is addressing the cold-start problem. When launching a new personalized feature, there is often no historical data to draw upon. FSPO provides a robust initial data curation pipeline that can deliver immediate value, as evidenced in the tasks studied in this work. This extends to situations where an organization has a wealth of user data, but not in a format amenable to LLMs (e.g, a housing and neighborhood commerce network such as BILT, which has a set of user transaction patterns and platform engagement not standard to LLMs). In such instances, a synthetic preference dataset can be designed using FSPO, based on the existing data signals. Additionally, as real preference data is collected, it can be integrated with or used to fine-tune the synthetically trained model, demonstrating how FSPO can serve as a critical foundation and accelerator. Furthermore, works such as AlpacaFarm have been introduced for prototyping/development of preference-based systems. As stated in their abstract, synthetic data such as LLM Prompts can simulate human feedback that is 50x cheaper than crowdworkers and display high agreement with humans (corroborated with our human study). Thus, in many real-world applications, this synthetic data generation pipeline can be used for benchmarking purposes that emulate a more realistic downstream application in personalization.

\textbf{2. Privacy-Sensitive Settings} Additionally, there exist applications where collecting and storing extensive user data is either impractical or undesirable due to privacy concerns. Consider an on-device AI assistant, a confidential workplace tool, financial/banking assistants, or a medical assistant. Here, approaches from the synthetic data pipeline, such as iterative persona construction (Figure 3, right) can be an appropriate approach to synthetically generate a user profile from the user preferences to elicit personalization, without needing a persistent user-written profile. This can additionally be constrained/controlled to not include any personally identifiable information from the preferences that are collected, which is advantageous, for example, in medical domains to avoid infringing on HIPAA. Collecting a comprehensive, detailed user profile is often intractable and inadvisable in such applications, but is beneficial for fine-tuning a personalized model, which our approach provides a controllable solution for. Similarly responses from a user may be difficult to collect in this instantiation as well. Here, our diverse response generation strategy may be a good fit, such as viewpoint conditioned responses, where viewpoints can be supervised by experts in a domain like medical professionals.

\textbf{3. Guided Data Curation \& Metrics} Finally, our synthetic data pipeline is instantiated on the guiding principles of structure and diversity, theoretically motivated by task-generation in meta-learning, which are readily transferable to real tasks. To ground this in a real problem, let’s consider the education domain that you have suggested, where student data might be available. Our approach can provide guidance on data curation or data selection for a personalized system in this domain. To concretely measure these principles, we study and empirically evaluate metrics that characterize the principle. For example, for diversity, we study the embedding similarity of responses as seen in Table 1, and we introduce a disagreement metric as seen in Figure 5 to capture the diversity of responses and users. This can be readily used to gauge the diversity of real preference data, such as capturing the diversity of education backgrounds of students or the diversity in tutoring conversations in an education setting. We characterize structure in preferences by the binary Shannon entropy of the preference labels, a metric we study in section 6. This can be used to identify underspecification of a user’s education background with the prompts and responses that they label preferences for, and potentially be used to filter users in the dataset that may be too noisy. Having inconsistencies in preferences or the user description makes the learning problem much more difficult, as described in prior work such as C-DPO and IPO. Overall, these metrics can guide the data selection/curation process of existing human data in domains such as education and has been validated on users in our human study, which indicates the effectiveness of the data curation approach.

To sum up, FSPO can be readily incorporated into several real world applications, where it can help provide a warm-start in data limited regimes or a dataset to prototype with, with settings where user-privacy is paramount, and additionally guide the curation and selection of human data, in applications such as education through the guiding principles of structure and diversity with the metrics proposed. 

\subsection{The Use of Large Language Models (LLMs)}

Large Language Models are used to assist with proofreading and minor wording improvements. All research ideas, experiments, and conclusions were conceived and validated by the authors. Additionally, tools such as Cursor were utilized as coding assistants during the development of the coding infrastructure for the project.

\end{document}